\documentclass[12pt]{article}

\usepackage[noend]{algpseudocode}
\usepackage{algorithm}
\usepackage{algorithmicx}
\usepackage{lmodern}
\usepackage{etoc}
\usepackage{scicite}
\usepackage{amsmath}
\usepackage{amsfonts}
\usepackage{mathtools}
\usepackage{scalerel}
\usepackage{amssymb}
\usepackage{bm}
\usepackage{makecell}
\usepackage{siunitx}
\usepackage[dvipsnames]{xcolor}
\usepackage{color,soul}
\usepackage{moreverb} 
\usepackage[normalem]{ulem}
\usepackage{standalone}
\usepackage{circuitikz}
\usepackage{hyperref}
\usepackage{tikz}
\usepackage{tabularx}
\usepackage{booktabs}
\usetikzlibrary{external}
\usepackage{physics}
\DeclareMathOperator*{\argmin}{arg\min}
\DeclareMathOperator*{\LO}{LO}
\DeclareMathOperator*{\GO}{GO}

\tikzset{
  font={\fontsize{9pt}{12}\selectfont}}

\newcolumntype{Y}{>{\centering\arraybackslash}X}
\newcolumntype{s}{>{\hsize=0.7\hsize}Y}

\usepackage{pgfplots}
\usepgfplotslibrary{fillbetween,colormaps,groupplots}

\pgfplotsset{hide scale/.style={
/pgfplots/xtick scale label code/.code={},
/pgfplots/ytick scale label code/.code={}}}

\pgfplotsset{
        table/search path={figures/},
    }

\pgfplotsset{
    colormap/gy/.style={
        colormap={gy}{[1pt]
            rgb(0pt)=(1,0.95,0.95);
            rgb(17pt)=(0.5, 0.5, 0.5);
            rgb(25pt)=(0.4, 0.4, 0.4);
            rgb(100pt)=(0.4, 0.4, 0.4);
        }
    }
}

\pgfplotsset{
    colormap/gyn/.style={
        colormap={gny}{[1pt]
            rgb(0pt)=(1,0.95,0.95);
            rgb(66pt)=(0.5, 0.5, 0.5);
            rgb(100pt)=(0.4, 0.4, 0.4);
        }
    }
}

\pgfplotsset{
    colormap/bw/.style={
        colormap={bw}{[1pt]
            rgb(0pt)=(0.1, 0.1, 0.1);
            rgb(25pt)=(0.1, 0.1, 0.1);
            rgb(50pt)=(0.25, 0.25, 0.25);
            rgb(100pt)=(1,0.95,0.95);
        }
    }
}

\pgfplotsset{
    /pgfplots/layers/Bowpark/.define layer set={
        axis background,axis grid,main,axis ticks,axis lines,axis tick labels,
        axis descriptions,axis foreground
    }{/pgfplots/layers/standard},
}

\usepackage{caption}

\graphicspath{{./figures/}{./tables/}}

\makeatletter
\def\input@path{{./figures/}{./tables/}}
\makeatother

\usepackage{times}
\usepackage[]{url}

\DeclareMathOperator*{\Aop}{\scalerel*{\mathbb{A}}{\sum}}

\newcommand{\appropto}{\mathrel{\vcenter{
  \offinterlineskip\halign{\hfil$##$\cr
    \propto\cr\noalign{\kern2pt}\sim\cr\noalign{\kern-2pt}}}}}

\makeatletter
\renewcommand{\thesubsection}{%
  \ifnum\c@section<1 \@arabic\c@subsection
  \else \thesection.\@arabic\c@subsection
  \fi
}
\makeatother

\newcounter{lastnote}

\title{{A machine learning approach for fighting the curse of dimensionality in global optimization}} 

\author
{Julian F. Schumann,$^{1}$ Alejandro M. Arag\'{o}n$^{1\ast}$\\
\\
\normalsize{$^{1}$Department Precision and Microsystems Engineering, Technological University Delft,}\\
\normalsize{Mekelweg 2, 2628 CD Delft, Netherlands}\\
\\
\normalsize{$^\ast$To whom correspondence should be addressed; E-mail:  a.m.aragon@tudelft.nl }
}

\date{}

\newcommand*{\algrule}[1][\algorithmicindent]{\hspace*{.5em}\vrule\vrule
width 0pt height \baselineskip depth .25\baselineskip\hspace*{\dimexpr#1-.5em}}

\makeatletter
\newcount\ALG@printindent@tempcnta
\def\ALG@printindent{%
    \ifnum \theALG@nested>0
    \ifx\ALG@text\ALG@x@notext
    \else
    \unskip
    \ALG@printindent@tempcnta=1
    \loop
    \algrule[\csname ALG@ind@\the\ALG@printindent@tempcnta\endcsname]%
    \advance \ALG@printindent@tempcnta 1
    \ifnum \ALG@printindent@tempcnta<\numexpr\theALG@nested+1\relax
    \repeat
    \fi
    \fi
}%
\usepackage{etoolbox}
\patchcmd{\ALG@doentity}{\noindent\hskip\ALG@tlm}{\ALG@printindent}{}{\errmessage{failed to patch}}
\makeatother

\AtBeginEnvironment{algorithmic}{\lineskip0pt}

\usepackage[scaled]{beramono}
\usepackage[T1]{fontenc}

\algtext*{EndWhile}
\algtext*{EndFor}
\algtext*{EndIf}
\algdef{SE}[DOWHILE]{Do}{doWhile}{\algorithmicdo}[1]{\algorithmicwhile\ #1}%

\makeatletter
\renewcommand{\ALG@beginalgorithmic}
{\tt}
\makeatother

\algnewcommand\algorithmicto{\textbf{to}}
\algrenewtext{While}{\textbf{while}\ }
\algrenewtext{Procedure}[2]{\textbf{function}\ \textproc{#1}\ifthenelse{\equal{#2}{}}{}{(#2)}}

\algrenewcommand\algorithmicrequire{\textbf{Input:}}
\algrenewcommand\algorithmicensure{\textbf{Output:}}

\usepackage{eqparbox}
\algrenewcommand{\algorithmiccomment}[1]{\hfill\eqparbox{COMMENT}{\color{gray} \it-- #1}}
\newcommand{\inlinecomment}[1]{{\color{gray} \it-- #1}}

\begin{document}


\immediate\write18{texcount -nc -inc -sum \jobname.tex > wordcount.tex}
\newcommand\wordcount{\verbatiminput{wordcount.tex}}


\immediate\write18{texcount -char -inc -freq \jobname.tex > charcount.tex}
\newcommand\charcount{\verbatiminput{charcount.tex}}


\baselineskip24pt


\maketitle


\begin{abstract}
Finding global optima in high-dimensional optimization problems is extremely challenging since the number of function evaluations required to sufficiently explore the search space increases exponentially with its dimensionality. Furthermore, multimodal cost functions render local gradient-based search techniques ineffective.
To overcome these difficulties, {we propose to trim uninteresting regions of the search space where global optima are unlikely to be found by means of autoencoders, exploiting the lower intrinsic dimensionality of certain cost functions; optima are then searched over lower-dimensional latent spaces.}
The methodology is tested on benchmark functions and on {multiple variations of a structural topology optimization problem, where we show that we can estimate this intrinsic lower dimensionality and based thereon obtain the global optimum at best or superior results compared to established optimization procedures at worst.}
\end{abstract}


\section*{Introduction}
Ever since Newton minimized the resistance of a radial symmetric body in a fluid flow~\cite{Newton}, countless optimization problems have occupied the minds of scholars. Today, as humanity struggles with an ever-increasing population and a finite amount of resources~\cite{Thogersen2014}, optimization plays a fundamental role in enabling greater efficiency in the use of materials and energy~\cite{Cho2016,Zhau2017,Jiminez2019,Xiong2020}.  The advent of computers and the increase in computational power over the years \cite{Shalf2020} has also enabled the solution of increasingly complex optimization problems in numerous fields such as engineering, medicine, and economics~\cite{Long2019, Pillardy2001,Chang2011}. Still, the optimization of high-dimensional problems that are characterized by a non-convex ad multimodal cost function---where the global optimum sought is one of many local optima---still poses significant challenges. This is caused by the so-called ``curse of dimensionality,'' whereby the number of cost function evaluations required for a sufficiently thorough survey of the search space increases exponentially with the number of dimensions~\cite{Chen2015,Guirguis2020}. This becomes even more troublesome for computationally-intensive problems, where the cost of function evaluations further limits the size and consequently the dimensionality of the space that can be explored.

A plethora of algorithms have been proposed over the years to optimize high-dimensional problems~\cite{Molina2018,Mauvcec2019}.
They are, however, all limited by the so-called ``no free lunch theorem,'' whereby optimization algorithms theoretically perform the same when averaged over all possible optimization problems~\cite{Wolpert1997,Adam2019}. As a result, a certain algorithm can only be more efficient than others on a certain set of optimization problems, where it is best at exploiting a common property of these problems' cost functions.
For example, gradient-based algorithms are superior on unimodal, differentiable cost functions, but often are unable to find global optima on multimodal and non-convex cost functions.
When trying to overcome this limitation and improve over gradient-based search, one usually turns towards global optimization algorithms,\footnote{Here \textit{global} does not imply surveying the entire space, but just different portions of it in a way that the search can escape local optima basins~\cite{Jones1998,Storn1997}.} for instance surrogate model-based Bayesian optimization or metaheuristic search~\cite{Zhig2021}. On the one hand, Bayesian optimization was \textit{designed} for problems with expensive cost function evaluations~\cite{Jones1998}, but struggles with high-dimensional search spaces, with recent works considering problems with at most a few hundred dimensions~\cite{Amine2018,Erikson2019}. This is because the construction of the surrogate model, which requires the repeated inversion of a dense, often ill-conditioned (near singular), and large matrix~\cite{Zhig2021}---with size proportional to the dimensionality of the problem~\cite{Jones1998}---is not only time-consuming but can also lead to memory issues~\cite{Cosme2018}. On the other hand, metaheuristic search algorithms can be used for high-dimensional problems, but are only applicable to relatively inexpensive cost functions due to the vast number of function evaluations required. Population-based search algorithms like genetic algorithms~\cite{Mitchell1998}, differential evolution~\cite{Storn1997}, or particle swarm optimization~\cite{Kennedy1995} are widely used examples of heuristic search methods.

In this work we focus on problems where large partions of the search space can be discarded.
Model order reduction (MOR) can then used to limit the search space to regions where the global optimum is likely to be found. {One possible approach would be to use a MOR based on random linear embeddings, where a linear mapping is used between search and latent spaces, chosen to minimize the loss of information in the training set~\cite{wang_bayesian_2016}. But in this work, we will use autoencoders instead of linear MOR (such a linear embeddings or proper orthogonal decomposition), as especially for non-linear MOR problems, they have been shown to be superior~\cite{Kashima2016,Hartman2017,Agostino2018,Lee2019,Kutnyiok2019,Maulik2020}.} These autoencoders then allow us to obtain a \textit{latent space} with drastically reduced dimension, by using two feed-forward neural networks that learn to copy the input to the output: The encoder network maps the input into the lower-dimensional latent space, and the decoder network is trained to reconstruct the input~\cite{hinton2006reducing}. Once an autoencoder is trained, global optimization over the latent space then becomes feasible. Since global optimization over lower-dimensional spaces is covered extensively in the literature, be it by heuristic search for inexpensive cost functions \cite{Floudas2009,Kora2017, Dabbagh2018, wang2018particle} or Bayesian optimization for comparatively expensive cost functions \cite{Jones1998,Snoek2012,Shahriari2015}, the main focus of this work is on autoencoders.

The use of autoencoders to enable global optimization in latent spaces was first studied by Costa in a constrained wind-hydro coordination problem~\cite{Costa2008}, although it was only applied to a relatively low-dimensional problem (\num{120} dimensions with a reduction in dimensionality by only a factor of 2). A similar work was also pursued by Miranda \textit{et al.}~\cite{Miranda2014}, who also used this method to achieve faster convergence on a number of benchmark functions. Gao \textit{et al.}~applied autoencoders to fit a geology model to data, reducing the dimensionality from \num{38400} dimensions down to \num{140}~\cite{Gao2019geo,Gao2020geo}. Eismann \textit{et al.}~\cite{Eismann2017} used an autoencoder in a fluid resistance minimization problem, reducing the \num{9408}-dimensional problem to a \num{20}-dimensional latent space. They also added a neural network-based surrogate model that mapped a latent space sample to a scalar value. Training this network and the autoencoder simultaneously, so that the surrogate network's output approximates the cost function value of the high-dimensional encoder input, led to a more \textit{structured latent space} with more evenly spaced local optima and more moderate gradients; this was beneficial for using Bayesian optimization over the latent space. Although improved designs for the drag minimization problem were obtained, the procedure used is flawed because training samples for the autoencoder were generated randomly and therefore its effectiveness is limited, since the training samples have no inherent features the autoencoder could learn. {Similarly, Grosnit \textit{et al.} could also improve their final optimized result by adding further objectives to the loss function during the autoencoder's training, but instead of training a surrogate model to separate fit and unfit samples in latent space, they instead added a loss term that punishes large distances between selected samples with similar fitness and short distances between samples with large differences in fitness~\cite{grosnit_high-dimensional_2021}.} Meanwhile, Kudyshev \textit{et al.}~proposed the use of an adversarial autoencoder to enforce a certain distribution of designs in latent space when optimizing phononic metamaterials~\cite{Kudyshev2020}. {Tripp \textit{et al.}~found an improvement in results by repeatedly retraining an autoencoder with an expanded training set---including newly found points form previous iterations of the optimization over latent space---and putting a higher weight on the fitter training samples in the autoencoder loss function~\cite{tripp_sample-efcient_2020}.} Autoencoders for reducing the search space have also been employed in other works, for example in the design of electromagnetic circuits~\cite{Tucci2019,Barmada2019,tucci_regularized_2021}, optical microstructures~\cite{Yang2018}, mechanical structures such as springs \cite{Tutum2018} or wheels \cite{Oh2019}, {or the activation maximization of already trained neural networks, where one tries to find the input of a neural network that maximizes the activation of specific neurons further down the network~\cite{ponce_evolving_2019,xiao_gradient-free_2020,verma_uncertainty-aware_2021}. Another application of using autoencoders to optimize over latent space can be found in the recovery of incomplete trajectories of specific human joints~\cite{lohit_recovering_2021}.}

{Similar concepts---\textit{i.e.,} the use of autoencoders to create a latent space over which to optimize---are also commonly employed to solve high dimensional, but discrete problems, such as the design of chemical molecules~\cite{kusner_grammar_2017, gomez-bombarelli_automatic_2018, griffiths_constrained_2020, tripp_sample-efcient_2020, siivola_good_2021,mendez-lucio_geometric_2021,yao_inverse_2021, zhang_-depth_2021, grosnit_high-dimensional_2021} and symbolic regression~\cite{kusner_grammar_2017, tripp_sample-efcient_2020}. But here the main purpose for the use of the autoencoders lies in generating a continuous space where common optimization algorithms such as Bayesian optimization can be used, while training samples for the autoencoder are mostly drawn from existing data sets.}

While the aforementioned works already use autoencoders to reduce the dimensionality of an optimization problem, there is lack of justification for their successful application in regard to the ``no free lunch theorem''---besides reporting improved designs. For example, none of the aforementioned works explains why their training data has a lower intrinsic dimensionality which the autoencoder might exploit, and some even use completely random training samples~\cite{Eismann2017}. Furthermore, the authors generally do not explain why their proposed methods should still be able to find global optima reliably despite the reduction in dimensionality, which makes the use of autoencoders for global optimization algorithms questionable at best {(although exception exists~\cite{wang_bayesian_2016}). Additionally, in most problems no considerations are made about the required size of the autoencoder (unless the intrinsic dimensionality can be clearly extracted from the problem formulation as done by Tucci \textit{et al.}~\cite{tucci_regularized_2021}), which further undermines the validity of this approach. Consequently, it is unclear for what kind of other optimization problems this approach could be used. Lastly, most works rely either on already existing data sets to get training data, or the data generation is only valid for a specific optimization problem (for example by relying on symmetry~\cite{Kudyshev2020}), which makes the generalization of their methods to other problems difficult, if not impossible. Therefore, our contribution is the provision and justification of a general framework for using autoencoders in pursuit of global minima for any possible problem fulfilling the underlying assumption that large parts of the search space can be discarded. This includes methods for building the training set for the autoencoder or determining the best dimensionality of the latent space it creates.}

Exploring sufficiently the search space in pursuit of a global optimum is limited by two main factors, namely the dimensionality $n$ of the search space and the time required for a single cost function evaluation $t_c$. In this work we propose a generalized framework for autoencoder enabled global optimization that addresses the former, enabling a significantly faster convergence on some specific optimization problems, where the objective is to solve
\begin{equation}
    \bm{x}_{\min}= \underset{\bm{x}\in X}{\text{argmin}} \; c(\bm{x}), \;\; X\subset \mathbb{R}^n \label{eq:stand_opt_problem}
\end{equation} 
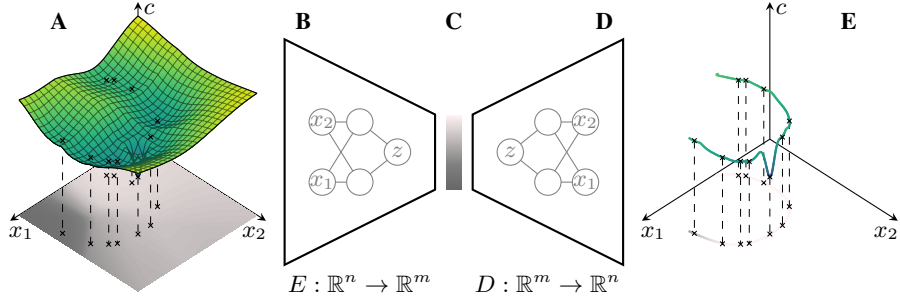
\begin{figure}
\centering
\tikzsetnextfilename{Figure_1}
\begin{tikzpicture}[]

\begin{axis}[at={(0.1cm,-0.5cm)},
width=3.4cm,
height=4cm,
view={135}{40},
	    scale only axis=true,
            enlargelimits=false,
	    axis lines=center,
            xmin=0,
            xmax=2.15,
            xtick={\empty},
            ymin=0,
            ymax=2.15,
            colormap/gy,
            ytick={\empty},
	 zmin = 0,
	 zmax = 6,
            ztick={\empty},	
            xmajorgrids=true,
            ymajorgrids=true,
            legend style={fill=white, fill opacity=0.5, draw opacity=1, text opacity=1},
            legend pos= north east,
            legend cell align=left]

	\addplot3[
            mesh/rows=101,
            mesh/cols=101,
            surf,  
            colormap/gy,  
            shader = interp,	
	 point meta = \thisrow{z},
            point meta min= 1.5,   
            point meta max=  5.5,
            ] table[x=x1, y=x2, z=n] from {Figure_1/X.txt};

	\addplot3[dashed, very thin,
            black,mark=x, mark size=1.25pt,mark options={solid,thin}
            ] table[x=x1, y=x2, z=c] from {Figure_1/X0.txt};

	\addplot3[
            mesh/rows=101,
            mesh/cols=101,
            surf,  
	 shader= interp,
            colormap/viridis,  
            point meta min= 1.5,   
            point meta max=  5.5,
            ] table[x=x1, y=x2, z=c] from {Figure_1/X.txt};
	\addplot3 [very thin, black, opacity=0.5,
            ] table[x=x1, y=x2, z=c] from {Figure_1/X_mesh.txt};
	\addplot3 [thin, black,
            ] table[x=x1, y=x2, z=c] from {Figure_1/X_mesh_2.txt};

\end{axis}

\begin{axis}[at={(6.2cm,2cm)},
hide axis,
    scale only axis,
    height=0pt,
    width=0pt,
    colormap/gyn,
    colorbar horizontal,
    point meta min=0,
    point meta max=1,
    colorbar style={
        width=1cm,
        rotate=-90,
        xtick={\empty},
axis line style={draw=none},
    },
 colorbar/width = 0.2cm,
    ]
    \addplot [draw=none] coordinates {(0,0)};

\end{axis}

\begin{axis}[at={(8.5cm,-0.5cm)},
width=3.4cm,
height=4cm,
view={135}{40},
	       scale only axis=true,
            enlargelimits=false,
	       axis lines=center,
            xmin=0,
            xmax=2.15,
            xtick={\empty},
            ymin=0,
            ymax=2.15,
            colormap/gy,
            ytick={\empty},
	 zmin = 0,
	 zmax = 6,
            ztick={\empty},	
            xmajorgrids=true,
            ymajorgrids=true,
            legend style={fill=white, fill opacity=0.5, draw opacity=1, text opacity=1},
            legend pos= north east,
            legend cell align=left,
            ]

	\addplot3+[mesh,thin,mark=none,
	 shader= interp,
            colormap/gy, thick, 
	 point meta = \thisrow{z},
            point meta min= 1.5,   
            point meta max=  5.5,
            ] table[x=x1, y=x2, z=n] from {Figure_1/Z1.txt};
\addplot3+[mesh,thin,mark=none,
	 shader= interp,
            colormap/gy, thick, 
	 point meta = \thisrow{z},
            point meta min= 1.5,   
            point meta max=  5.5,
            ] table[x=x1, y=x2, z=n] from {Figure_1/Z2.txt};

	\addplot3[dashed, very thin,
            black,mark=x, mark size=1.25pt,mark options={solid,thin}
            ] table[x=x1, y=x2, z=c] from {Figure_1/X0.txt};

           \addplot3+[mesh,thin,mark=none,
            colormap/viridis, thick,
	 shader= interp,
	 point meta = z, 
            point meta min= 1.5,   
            point meta max=  5.5,
            ] table[x=x1, y=x2, z=c] from {Figure_1/Z1.txt};

           \addplot3+[mesh,thin,mark=none,
            colormap/viridis, thick,
	 shader= interp,
	 point meta = z, 
            point meta min= 1.5,   
            point meta max=  5.5,
            ] table[x=x1, y=x2, z=c] from {Figure_1/Z2.txt};

\end{axis}

	\draw [black, thick] (5.75,1)--(5.75,2)--(3.75,3)--(3.75,0)--cycle;
	\draw [black, thick] (6.25,1)--(6.25,2)--(8.25,3)--(8.25,0)--cycle;

	\draw [black!50!white](4.25,1.1) -- (4.75,1.1);
	\draw [black!50!white](4.25,1.1) -- (4.75,1.9);
	\draw [black!50!white](4.25,1.9) -- (4.75,1.9);
	\draw [black!50!white](4.25,1.9) -- (4.75,1.1);
	\draw [black!50!white](4.75,1.1) -- (5.25,1.5);
	\draw [black!50!white](4.75,1.9) -- (5.25,1.5);
	\fill [draw=black!50!white, fill=white] (4.25,1.1) circle(0.175) node[black!50!white] {\noexpand\noexpand\noexpand\footnotesize{$x_1$}};
	\fill [draw=black!50!white, fill=white] (4.25,1.9) circle(0.175) node[black!50!white] {\noexpand\noexpand\noexpand\footnotesize{$x_2$}};
	\fill [draw=black!50!white, fill=white] (4.75,1.1) circle(0.175);
	\fill [draw=black!50!white, fill=white] (4.75,1.9) circle(0.175);
	\fill [draw=black!50!white, fill=white] (5.25,1.5) circle(0.175) node[black!50!white] {\noexpand\noexpand\noexpand\footnotesize{$z$}};

	\draw [black!50!white](7.75,1.1) -- (7.25,1.1);
	\draw [black!50!white](7.75,1.1) -- (7.25,1.9);
	\draw [black!50!white](7.75,1.9) -- (7.25,1.9);
	\draw [black!50!white](7.75,1.9) -- (7.25,1.1);
	\draw [black!50!white](7.25,1.1) -- (6.75,1.5);
	\draw [black!50!white](7.25,1.9) -- (6.75,1.5);
	\fill [draw=black!50!white, fill=white] (7.75,1.1) circle(0.175) node[black!50!white] {\noexpand\noexpand\noexpand\footnotesize{$x_1$}};
	\fill [draw=black!50!white, fill=white] (7.75,1.9) circle(0.175) node[black!50!white] {\noexpand\noexpand\noexpand\footnotesize{$x_2$}};
	\fill [draw=black!50!white, fill=white] (7.25,1.1) circle(0.175);
	\fill [draw=black!50!white, fill=white] (7.25,1.9) circle(0.175);
	\fill [draw=black!50!white, fill=white] (6.75,1.5) circle(0.175) node[black!50!white] {\noexpand\noexpand\noexpand\footnotesize{$z$}};

	\node at (4.75,-0.25) {$E: \mathbb{R}^n \rightarrow \mathbb{R}^m$};
	\node at (7.25,-0.25) {$D: \mathbb{R}^m \rightarrow \mathbb{R}^n$};

\node at (0.25,0.425) {$x_1$};
\node at (3.35,0.425) {$x_2$};
\node at (1.95,3.4) {$c$};

\node at (8.65,0.425) {$x_1$};
\node at (11.75,0.425) {$x_2$};
\node at (10.35,3.4) {$c$};

\node at (0.75,3.25) {\textbf{A}};
\node at (11.25,3.25) {\textbf{E}};
\node at (4,3.25) {\textbf{B}};
\node at (6,3.25) {\textbf{C}};
\node at (8,3.25) {\textbf{D}};

\draw [white] (0,-0.75) -- (0,4) -- (12,4)--(12,-0.75) --cycle;

\end{tikzpicture}
\caption{Visualization of the underlying idea. Instead of optimizing over the whole domain $X\subset \mathbb{R}^n$ (\textbf{A}), an autoencoder is trained first. Here the encoder $E$ (\textbf{B}, the grayscale coding in \textbf{A} and \textbf{C} is used to represent this function) maps a sample $\bm{x}$ from the search space into the latent space $Z\subset \mathbb{R}^m$ (\textbf{C}). A decoder $D$ (\textbf{D}, the same grayscale coding is used, here in \textbf{C} and \textbf{E}) then transforms the latent space sample $\bm{z}$ back into higher-dimensional domain $X$, although the decoded samples will be part of the lower dimensional manifold $X_Z\subset X$ (\textbf{E}), over which the global optimization is then performed.}
\label{fig:schematic}
\end{figure}
in a limited amount of time.
To find the global minimum $\bm{x}_{\min}$, we assume that a lower-dimensional manifold $X_Z \subset X \subset \mathbb{R}^n$---which is the \textit{decoded latent space} of an autoencoder trained on a number of sampled local minima; the latter are found by randomly sampling the high-dimensional search space $X$ and subsequent local gradient-based search---goes through the basin $X_{\min}$ of the global optimum. This discards large uninteresting regions of the search space where we assume that the global optimum is unlikely to be found.
The procedure is schematically shown in \textbf{Fig.~\ref{fig:schematic}}, where $c(\bm{x})$ is the cost function of the high-dimensional space $X$ (in \textbf{Fig.~\ref{fig:schematic}A} $X \subset \mathbb{R}^2$ for visualization purposes, but the dimensionality could be arbitrarily high).
The encoder component $E$ of the autoencoder (\textbf{Fig.~\ref{fig:schematic}B}) maps a point $\bm{x} \in X$ to a lower-dimensional point $E (\bm{x}) = \bm{z} \in Z$ (\textbf{Fig.~\ref{fig:schematic}C}, where $z \in \mathbb{R}$). The decoder component $D$ (\textbf{Fig.~\ref{fig:schematic}D}) then maps such latent space samples back into the high-dimensional space, \textit{i.e.}, $D \left( \bm{z} \right)  \in X_Z $ (\textbf{Fig.~\ref{fig:schematic}E}). The specific $X_Z$ is then the result of the autoencoder training, where the autoencoder tries to minimize the error between the input sampled local minima (see $\times$ markers on $c(\bm{x})$ in \textbf{Fig.~\ref{fig:schematic}A}) and their reconstructed counterparts.
This proposed method is then viable if the latent space dimensionality $m$ is larger than or equal to the intrinsic dimensionality $m^*$ of the optimization problem---which can be understood as the number of features distinguishing usable regions of the search space (see Appendix~\ref{App:assumption})---and if $m^*$ is significantly smaller than $n$, then optimizing over $Z\subset \mathbb{R}^m$ would require significantly fewer cost function evaluations than optimizing over $X\subset\mathbb{R}^n$.
The procedure is described in more detail in \S~\hyperref[sec:methods]{\textbf{Methods}}.
It is worth mentioning that this method is only useful if random multi-start local search~\cite{Marti2003} is unable to find the global optimum, which is the method we use to generate the autoencoder's input data. Only then the optimization over latent space and the associated exploration of other promising regions of the high-dimensional search space could be effective, as the autoencoder is able to generate a decoded latent space $X_Z$ going through the basin of the global minimum even when trained with no samples in such basin.

\section*{Results} \label{sec:results}

The methodology is tested on constructed benchmark functions and on a topology optimization problem.

\subsection*{Benchmark functions} \label{sec:BM_test}
{
We first test the procedure on the following cost function
\begin{equation} \label{eq:c_1}
    c_1(\bm{x})= \left( {  \sum\limits_{i=2}^{1000}    1 + {c_{0,i}\over{\vert\vert \bm{x}-\bm{x}_i \vert\vert^2}} \over{ \sum\limits_{i=2}^{1000}  {1\over{\vert\vert \bm{x}-\bm{x}_i \vert\vert^2}}  }} + \underset{i\in \{2, \hdots, 1000\}}{\min}  \, 5\, \vert\vert \bm{x}-\bm{x}_i \vert\vert^2  \right)\left(  \min \left\{ 1, 4 \vert\vert \bm{x}-\bm{x}_1 \vert\vert^2 \right\}\right),
\end{equation}
where $\bm{x_i} \in \{\bm{x}_1,\hdots , \bm{x}_{\num{1000}}\}=\bm{X}$ are $\num{1000}$ points arrayed on a $5$-dimensional manifold ($M=5$) in the $n=\num{100}$ dimensional search space $X_1$. These points correspond to the local minima of $c_1$, with
\begin{equation}  
	c_1(\bm{x}_i)=\begin{cases} 0 & \text{for } i=1 \\ c_{0,i}, \in [1,2] & \text{otherwise}.  \end{cases}
\end{equation} }
The procedure is also tested on three other cost functions borrowed from the work of Abualigah \textit{et al.}~\cite{abualigah2021}:
\begin{equation}
    \begin{aligned}
    c_2(\bm{x})=& \sum\limits_{i=1}^n \left(-x_i \sin\left( \sqrt{\vert x_i\vert}\right) \right)+418.9829 n,\\
    c_3(\bm{x})=&{\pi\over{n}}\left( 10 \sin\left(\pi y(x_1)\right)^2 + \sum\limits_{i=1}^{n-1} \left(y(x_i)-1\right)^2 \left(1+ 10 \sin\left(\pi y(x_{i+1})\right)^2    +u (\bm{x})   \right)  \right)\\
    &y(x)={x+5\over{4}},\;\; u(\bm{x})=\sum\limits_{i=1}^n  100 \max\left\{0,\vert x_i\vert-10\right\}^4,\\
    c_4(\bm{x})=& 1+{1\over{\num{4000}}}\sum\limits_{i=1}^n\left( x_i^2\right)-\prod\limits_{i=1}^n\left( \cos\left({x_i\over{\sqrt{i}}}\right)\right).
    \end{aligned}
\end{equation}
Noteworthy, as these three functions' local minima are spread roughly evenly over the whole domain $X_i$, one can expect that the intrinsic dimensionality of the problem $m^*=n=100$, as there are no features in the training samples an autoencoder could learn. If this is indeed the case, the proposed method with $m< n$ should most likely not find the global optimum.

{
All of these four functions are minimized over an $n=100$ dimensional domain (with $X_1=[-1,1]^n$, $X_2=X_4=[-500,500]^n$, and $X_3=[-50,50]^n$), using our proposed method with latent spaces with varying dimensionality, \textit{i.e.}, $m = \left\{2,\hdots, 7 \right\}$. We also compare our results with those obtained by means of differential evolution, a commonly used global optimization algorithm.
For details on the creation of $\bm{X}$ and the implementation of the optimization see Appendix~\ref{App:POC1}.

\textbf{Fig.~\ref{fig:benchmark_results}} summarizes the results, where for each of the optimization runs, the lowest found cost function value after a certain number of cost function evaluations $n_F$ is shown; this is done for each of the steps of the proposed method, \textit{i.e.}, the generation of the autoencoder training samples (black lines), the optimization over latent space (thick colored lines), and finally the post processing (thin colored lines) (see \S~\hyperref[sec:methods]{\textbf{Methods}}).
\begin{figure}[!ht]
\centering
\tikzsetnextfilename{Figure_2}
\input{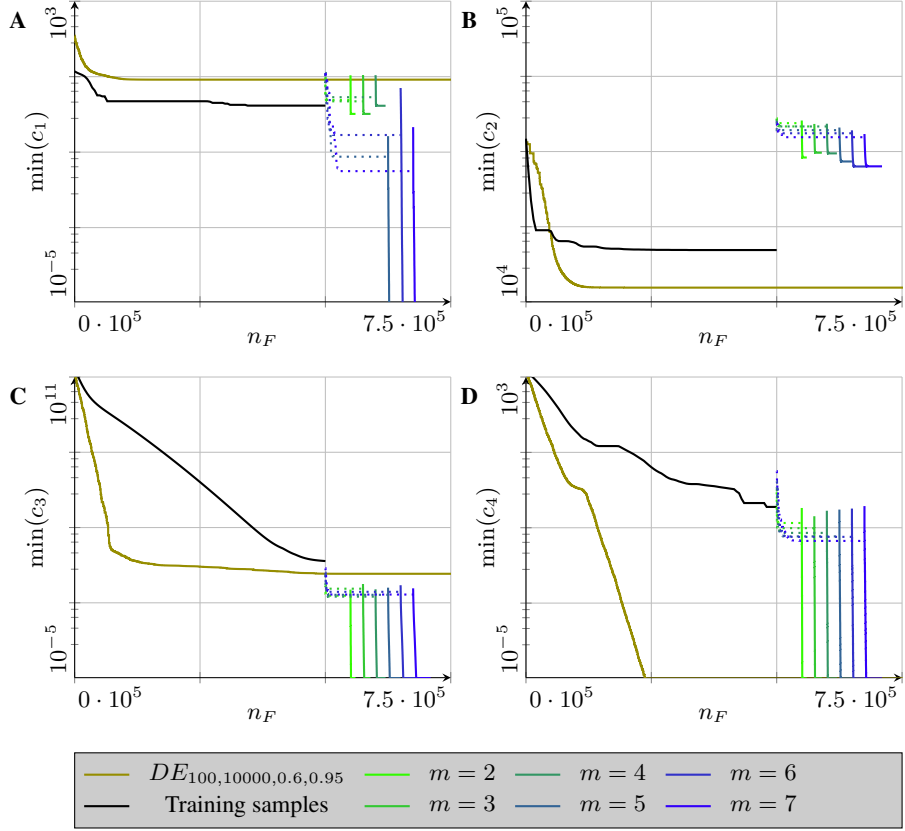}
\caption{{The results of optimizing the functions $c_1$ (\textbf{A}), $c_2$ (\textbf{B}), $c_3$ (\textbf{C}), and $c_4$ (\textbf{D}) for latent spaces of varying dimensionality $m$. The black lines show the lowest found function values after $n_F$ cost function evaluations during the first step of the proposed method, where $\lambda$ steps of local optimization are used on $\num{5000}$ random initial points to generate the training samples for the autoencoder. The colored dotted lines show the lowest found function values during the optimization over latent space after each cost function evaluation, while the colored solid lines show the same, except for the last step of the method, the post-processing. One can also see the results of using differential evolution in the high-dimensional search space in olive.}}
\label{fig:benchmark_results}
\end{figure}

It can be seen in \textbf{Fig.~\ref{fig:benchmark_results}A} that the global optimum of $c_1$ can only be found with the proposed method for $m\geq 5$, while both the proposed method with $m<5$ as well as differential evolution over the search space are unsuccessful. These results, which empirically indicate an intrinsic dimensionality $m^*=5$ for $c_1$ (as defined in Appendix \ref{App:assumption}), were expected (since one could predict $m^*=M$ for this case, see Appendix \ref{App:proof_c1}).

A similarly expected result could be found in the optimization of $c_2$ (see \textbf{Fig.~\ref{fig:benchmark_results}B}), where the optimization was indeed not successful, since $m\ll m^*=n$. }

In contrast, the optimization succeeds for $c_3$ and $c_4$ in spite of using an autoencoder with $m \ll n$, which empirically indicates that these problem have a far lower intrinsic dimensionality with $m^*\leq 2 \ll n$ ($n=100$) (see \textbf{Figs.~\ref{fig:benchmark_results}C} and \textbf{\ref{fig:benchmark_results}D}, respectively). The reason for this is that contrary to $c_2$, the global optimum of these functions lies precisely at the center of an evenly spread cluster of local minima. As is shown in Appendix \ref{App:proof_eq_mean}, in such cases the intrinsic dimensionality $m^*$ of the problem might fall as low as $m^*=1$, and therefore, the proposed method was indeed able to find the global optimum (although there were no parts of the search space we could discard \textit{a priori}).

\subsection*{Optimizing for fracture anisotropy in chocolate unit cells} \label{sec:CM}
\begin{figure}[!ht]
\centering
\tikzsetnextfilename{Figure_3}
\input{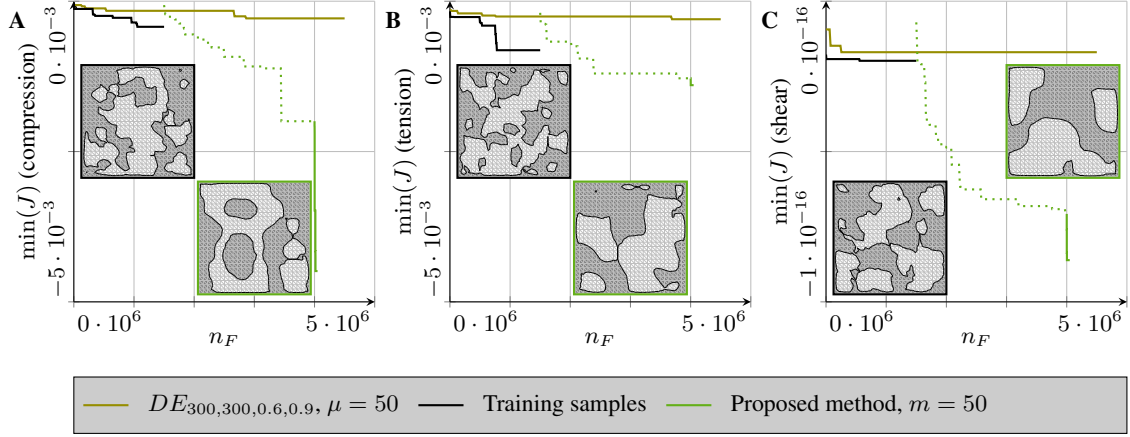}
\caption{{Results of running the optimization methodology for the  \hyperref[sec:CM]{\textbf{fracture anisotropy maximization}} problem (\textit{i.e.}, the cost function $J$ of Eq.~\eqref{eq:chocolate}). The three cases considered include compression \textbf{(A}), tension (\textbf{B}), and shear (\textbf{C}).
The different steps of the proposed method are marked as in \textbf{Fig. \ref{fig:benchmark_results}}. The best designs found after generating the training samples and after post processing are shown as well, respectively, with black and colored frames.}}
 \label{fig:compliance_min}
\end{figure}

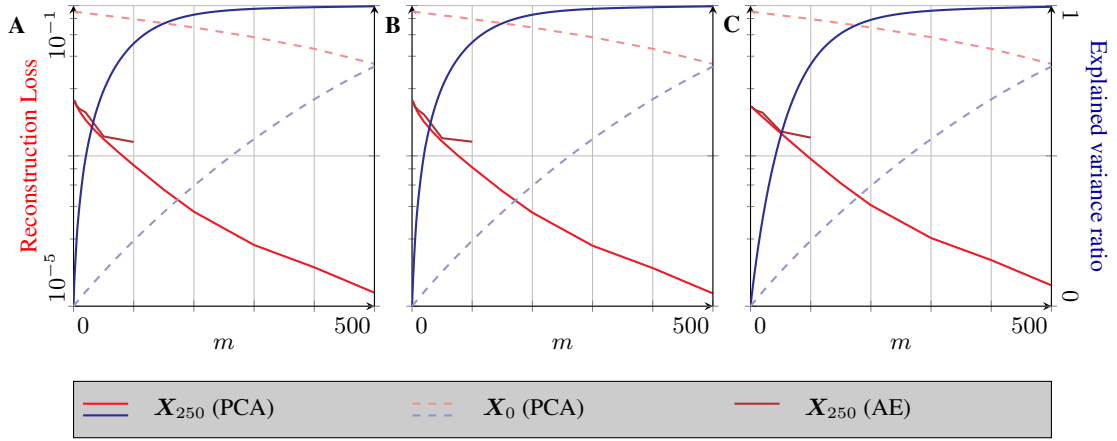
\begin{figure}[!ht]
\centering
\tikzsetnextfilename{Figure_14}
\begin{tikzpicture}[]
	
	\definecolor{color_r1}{RGB}{237, 27, 35} 			
	\definecolor{color_r2}{RGB}{246,141,146} 			
	\definecolor{color_r3}{RGB}{175,50,53} 			
	
	\definecolor{color_b1}{RGB}{45, 47, 146} 			
	\definecolor{color_b2}{RGB}{150,151,200}			

\begin{scope}[xshift = 0cm, yshift = 0cm]
	\begin{semilogyaxis}[
			at={(1cm,1cm)},
			width=4cm,
			height=4cm,
			scale only axis=true,
			axis lines = left,
			hide scale, 
			xmin=0,
			xmax=500,
			xtick={0, 100,200,300,400,500},
			xticklabels={,,,,,},
			scaled x ticks = false,
			ymin=0.00001,
			ymax=0.1,
			ytick={0.00001, 0.001, 0.1},
			yticklabels={,,,},
			minor ytick = 	{
						0.0784e-3 ,0.2116e-3, 0.4096e-3, 0.6724e-3,
						0.0784e-1, 0.2116e-1, 0.4096e-1, 0.6724e-1},
			yminorgrids=false,
			xmajorgrids=true,
			ymajorgrids=true
		]
	             
		\addplot[color = color_r1, solid, thick] 
		table[x =m, y=L]  {Figure_14/f_PCA_loss_2.txt};
		
		\addplot[color = color_r2, dashed, thick] 
		table[x =m, y=Lr]  {Figure_14/f_PCA_loss_2.txt};
		
		\addplot[color = color_r3, solid, thick] 
		table[x =m, y=L]  {Figure_14/f_AE_loss_2.txt};
	
	\end{semilogyaxis}
	         
	\begin{axis}[
			at={(1cm,1cm)},
			width=4cm,
			height=4cm,
			scale only axis=true,
			axis y line = right,
			axis x line = none,
			hide scale,
			xmin=0,
			xmax=500,
			xtick={},
			xticklabels={},
			scaled x ticks = false,
			ymin=0,
			ymax=1,
			ytick={0, 0.5, 1},
			yticklabels={,,},
			yminorgrids=false,
			xmajorgrids=false,
			ymajorgrids=false
		]
		
		\addplot[color = color_b1, solid, thick] 
		table[x =m, y=V]  {Figure_14/f_PCA_var_2.txt};
		
		\addplot[color = color_b2, dashed, thick] 
		table[x =m, y=Vr]  {Figure_14/f_PCA_var_2.txt};
		   
	\end{axis}
	         
	\node[black] at (0.25,4.75) {\textbf{A}};
	\node[black,right] at (0.9,0.75) {\noexpand\noexpand\noexpand\footnotesize{$0$}};
	\node[black,left] at (5.1,0.75) {\noexpand\noexpand\noexpand\footnotesize{$500$}};
	\node[black] at (3,0.5) {$m$};
	
	\node[black,right,rotate=90] at (0.75,0.9) {\noexpand\noexpand\noexpand\footnotesize{$10^{-5}$}};
	\node[black,left, rotate=90] at (0.75,5.1) {\noexpand\noexpand\noexpand\footnotesize{$10^{-1}$}};
	\node[black, rotate=90] at (0.4,3) {\textcolor{red}{Reconstruction Loss}};

\end{scope}

\begin{scope}[xshift =4.5cm, yshift = 0cm]
	\begin{semilogyaxis}[
			at={(1cm,1cm)},
			width=4cm,
			height=4cm,
			scale only axis=true,
			axis lines = left,
			hide scale, 
			xmin=0,
			xmax=500,
			xtick={0, 100,200,300,400,500},
			xticklabels={,,,,,},
			scaled x ticks = false,
			ymin=0.00001,
			ymax=0.1,
			ytick={0.00001, 0.001, 0.1},
			yticklabels={,,,},
			minor ytick = 	{
						0.0784e-3 ,0.2116e-3, 0.4096e-3, 0.6724e-3,
						0.0784e-1, 0.2116e-1, 0.4096e-1, 0.6724e-1},
			yminorgrids=false,
			xmajorgrids=true,
			ymajorgrids=true
		]
	             
		\addplot[color = color_r1, solid, thick] 
		table[x =m, y=L]  {Figure_14/f_PCA_loss_3.txt};
		
		\addplot[color = color_r2, dashed, thick] 
		table[x =m, y=Lr]  {Figure_14/f_PCA_loss_3.txt};
		
		\addplot[color = color_r3, solid, thick] 
		table[x =m, y=L]  {Figure_14/f_AE_loss_3.txt};
	
	\end{semilogyaxis}
	         
	\begin{axis}[
			at={(1cm,1cm)},
			width=4cm,
			height=4cm,
			scale only axis=true,
			axis y line = right,
			axis x line = none,
			hide scale,
			xmin=0,
			xmax=500,
			xtick={},
			xticklabels={},
			scaled x ticks = false,
			ymin=0,
			ymax=1,
			ytick={0, 0.5, 1},
			yticklabels={,,},
			yminorgrids=false,
			xmajorgrids=false,
			ymajorgrids=false
		]
		
		\addplot[color = color_b1, solid, thick] 
		table[x =m, y=V]  {Figure_14/f_PCA_var_3.txt};
		
		\addplot[color = color_b2, dashed, thick] 
		table[x =m, y=Vr]  {Figure_14/f_PCA_var_3.txt};
		   
	\end{axis}
	         
	\node[black] at (0.75,4.75) {\textbf{B}};
	\node[black,right] at (0.9,0.75) {\noexpand\noexpand\noexpand\footnotesize{$0$}};
	\node[black,left] at (5.1,0.75) {\noexpand\noexpand\noexpand\footnotesize{$500$}};
	\node[black] at (3,0.5) {$m$};

\end{scope}

\begin{scope}[xshift = 9cm, yshift = 0cm]
	\begin{semilogyaxis}[
			at={(1cm,1cm)},
			width=4cm,
			height=4cm,
			scale only axis=true,
			axis lines = left,
			hide scale, 
			xmin=0,
			xmax=500,
			xtick={0, 100,200,300,400,500},
			xticklabels={,,,,,},
			scaled x ticks = false,
			ymin=0.00001,
			ymax=0.1,
			ytick={0.00001, 0.001, 0.1},
			yticklabels={,,,},
			minor ytick = 	{
						0.0784e-3 ,0.2116e-3, 0.4096e-3, 0.6724e-3,
						0.0784e-1, 0.2116e-1, 0.4096e-1, 0.6724e-1},
			yminorgrids=false,
			xmajorgrids=true,
			ymajorgrids=true
		]
	             
		\addplot[color = color_r1, solid, thick] 
		table[x =m, y=L]  {Figure_14/f_PCA_loss_4.txt};
		
		\addplot[color = color_r2, dashed, thick] 
		table[x =m, y=Lr]  {Figure_14/f_PCA_loss_4.txt};
		
		\addplot[color = color_r3, solid, thick] 
		table[x =m, y=L]  {Figure_14/f_AE_loss_4.txt};
	
	\end{semilogyaxis}
	         
	\begin{axis}[
			at={(1cm,1cm)},
			width=4cm,
			height=4cm,
			scale only axis=true,
			axis y line = right,
			axis x line = none,
			hide scale,
			xmin=0,
			xmax=500,
			xtick={},
			xticklabels={},
			scaled x ticks = false,
			ymin=0,
			ymax=1,
			ytick={0, 0.5, 1},
			yticklabels={,,},
			yminorgrids=false,
			xmajorgrids=false,
			ymajorgrids=false
		]
		
		\addplot[color = color_b1, solid, thick] 
		table[x =m, y=V]  {Figure_14/f_PCA_var_4.txt};
		
		\addplot[color = color_b2, dashed, thick] 
		table[x =m, y=Vr]  {Figure_14/f_PCA_var_4.txt};
		   
	\end{axis}
	         
	\node[black] at (0.75,4.75) {\textbf{C}};
	\node[black,right] at (0.9,0.75) {\noexpand\noexpand\noexpand\footnotesize{$0$}};
	\node[black,left] at (5.1,0.75) {\noexpand\noexpand\noexpand\footnotesize{$500$}};
	\node[black] at (3,0.5) {$m$};
	 
	\node[black,left,rotate= - 90] at (5.25,0.9) {\noexpand\noexpand\noexpand\footnotesize{$0$}};
	\node[black,right, rotate= - 90] at (5.25,5.1) {\noexpand\noexpand\noexpand\footnotesize{$1$}};
	\node[black, rotate= - 90] at (5.6,3) {\textcolor{blue!60!black}{Explained variance ratio}};

\end{scope}

\filldraw[draw=black,fill=black!20!white] (1,-0.75) rectangle (14,0);
        \begin{axis}[
            at = {(1cm,-0.75cm)},
            axis lines = left,
            axis line style={draw=none},
            width=13cm,
            height=0.75cm,
	    scale only axis=true,
            enlargelimits=false,
            xlabel = ,
            ylabel = ,
            xmajorgrids=true,
	    ymajorgrids=true,
    	    legend columns = 3,
            legend style={fill=none, 
            			 cells = {align = left}, 
            			 draw=none, 
            			 at={(0,0.5)}, 
            			 anchor=west, 
            			 style={column sep=0.25cm, row sep = - 0.235cm}},
            legend cell align=center,
            xmin=0,
            xmax=0.1,
            xtick={\empty},
            ymin=0,
            ymax=0.1,
            ytick={\empty},
        ]
        
         \addplot[ 
                 color = color_r1,
                 solid,thick,
             ] coordinates {(1,1) (2,1)};

             \addplot[ 
                 color = color_r2,
                 solid,thick,dashed,
             ] coordinates {(1,1) (2,1)};
             
             \addplot[ 
                 color = color_r3,
                 solid,thick,
             ] coordinates {(1,1) (2,1)};
             
             \addlegendimage{empty legend};
             \addlegendimage{empty legend};
             \addlegendimage{empty legend};

             \addplot[ 
                 color = color_b1,
                 solid,thick,
             ] coordinates {(1,1) (2,1)};
             
             \addplot[ 
                 color = color_b2,
                 solid,thick,dashed,
             ] coordinates {(1,1) (2,1)};

             \addlegendimage{empty legend};

        \legend{\;,\;,\; ,
        {$\bm{X}_{250}$ (PCA) \textcolor{black!20!white}{000000000}}, 
        {$\bm{X}_{0}$ (PCA) \textcolor{black!20!white}{0000000000}},
        {$\bm{X}_{250}$ (AE)}, 
         \;,\;,\;};

        \end{axis}

 \end{tikzpicture}
 
\caption{{Clustering analysis of initial random samples $\bm{X}_{0}$ compared to locally optimized training samples for the autoencoder $\bm{X}_{250}$ for the \hyperref[sec:CM]{\textbf{fracture anisotropy maximization}} problem, for the cases of compression \textbf{(A}), tension (\textbf{B}), and shear (\textbf{C}). One can see the reconstruction loss and (if available) the explained variance ratio for different MOR-techniques and different latent space dimensionalities $m$, with principal component analysis being used for both cases, and autoencoders being used only for the training samples.}}
 \label{fig:rec_loss_comp}
\end{figure}

{
After testing the proposed method on benchmark functions, we turn our attention to a more complex optimization problem.
In that problem, we are interested in obtaining an optimized material distribution that maximizes fracture resistance anisotropy in a porous unit cell of chocolate.
To that end we use topology optimization (TO)~\cite{Bendsoe1989,Bendsoe1988,bendsoe2003}, which is a computational iterative procedure widely used in structural optimization that combines finite element analysis (FEA) with a gradient-based optimizer.
TO can therefore be used to find the most adequate material layout within a computational domain to optimize a given objective function subject to constraints.
Following the work of Souto \textit{et al.}~\cite{souto_edible_2022}, we subject a chocolate unit cell to two loading cases and we obtain the topology so that the resulting structure is very brittle for one of the loading cases and very tough for the other one.
The topology optimization problem is formally stated as
\begin{equation}
\begin{split}
\text{minimize} \quad  & J  = \omega J_1 - (1-\omega) J_2 \\
\text{such that} \quad & \boldsymbol K_i \boldsymbol U_i = \boldsymbol F_i, \quad i= \left\{ 1,2 \right\}, \\
 &  V = \bar{V},\\ \label{eq:chocolate}
\end{split}
\end{equation}where $\omega$ is a parameter that weights one loading case over the other (we set $\omega=0.5$ for equal weight), $J_i$ is an aggregation of the energy release rates $G_{j}$ computed at $N$ nodes along the boundary of the topology (chocolate-void interfaces):
\begin{equation}\label{eq:J}
J_i = \frac{1}{N}\sum_{j=1}^{N} G_{j},
\end{equation}
$\boldsymbol {K}_i \boldsymbol{U}_i = \boldsymbol{F}_i$ is the discrete form of equilibrium for the $i$th loading case, $V$ is the volume of chocolate and $\bar{V}$ the target volume (set here as $\bar{V} = 50\%$).
Energy release rates are computed using topological derivatives---borrowing concepts from linear elastic fracture mechanics for brittle solids---using an enriched finite element formulation. A level set function that is discretized by means of radial basis functions is used to describe the topology. For more details the reader is referred to Souto \textit{et al.}~\cite{souto_edible_2022} and Zhang \textit{et al.}~\cite{zhang2022tailoring} and references therein.

\begin{figure}[!ht]
\centering
\tikzsetnextfilename{Figure_15}
\input{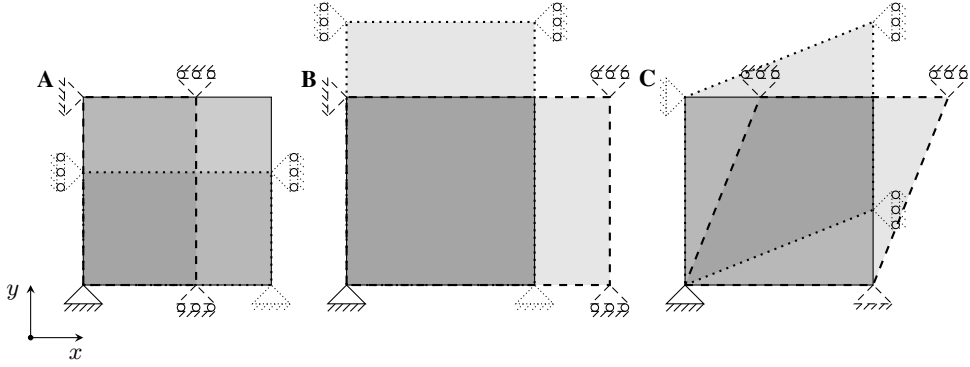}
\caption{{Three different cases are considered for the  \S~\hyperref[sec:CM]{\textbf{fracture anisotropy maximization}} problem, namely uniaxial compression (\textbf{A}), uniaxial tension (\textbf{B}), and shear (\textbf{C}). For each case, the dashed deformation corresponds to $J_1$ and the dotted deformation to $J_2$ respectively (see Equation \eqref{eq:chocolate}), which are prescribed by periodic boundary conditions (see equation \eqref{eq:periodic}). It has to be noted that the introduction of void elements into the unit cell can lead to curved boundaries, with identical curves on opposite edges of the unit cell.}}
\label{fig:load_cases}
\end{figure}

Three different optimization cases are considered, defined by different imposed deformations, namely uniaxial compression, uniaxial tension, and shearing. These are prescribed via periodic boundary conditions along the boundary of the unit cell $\Gamma$; given two points $\bm{x}_a, \bm{x}_b \in \Gamma$ located at opposite periodic edges of the unit cell (with $\bm{x}_b - \bm{x}_a$ being parallel to either the $x$ or $y$ axis), the periodic boundary condition states that
\begin{equation} \begin{aligned}
	&\bm{u}\left(\bm{x}_a\right) - \bm{u}\left(\bm{x}_b\right) = \bm{D}_i \left(\bm{x}_a - \bm{x}_b\right).
\label{eq:periodic}
\end{aligned}\end{equation}
 For the case of compression, one then uses the following $\bm{D}_i$:
\begin{equation}
	\bm{D}_1 = \begin{pmatrix} -0.1 & 0 \\ 0 & 0 \end{pmatrix}, \quad \bm{D}_2 = \begin{pmatrix} 0 & 0 \\ 0 &- 0.1 \end{pmatrix}.
\end{equation}
For the tension case, one uses 
\begin{equation}
	\bm{D}_1 = \begin{pmatrix} 0.1 & 0 \\ 0 & 0 \end{pmatrix}, \quad \bm{D}_2 = \begin{pmatrix} 0 & 0 \\ 0 & 0.1 \end{pmatrix}.
\end{equation}
Finally for the shearing case
\begin{equation}
	\bm{D}_1 = \begin{pmatrix} 0 & 0.1 \\ 0 & 0 \end{pmatrix}, \quad \bm{D}_2 = \begin{pmatrix} 0 & 0 \\ 0.1 & 0 \end{pmatrix} \, ,
\end{equation}
with the resulting unit cell shapes (assuming a completely solid square) sketched in \textbf{Fig. \ref{fig:load_cases}}. More details on prescribing periodic boundary conditions can be found in the work of Danielsson \textit{et al.}~\cite{danielsson2002three}.

These topology optimization problems are solved on an initial finite mesh composed of $n_y\times n_x=26 \times 26$ finite elements, where the values of the level set function on the outer nodes of the mesh is prescribed; the overall dimensionality of the problem is therefore $n=625$. This local topology optimization problem is then incorporated into the proposed method, where an adversarial autoencoder with $m=50$ dimensions is chosen, while a surrogate neural network is used during training (see Appendix \ref{App:DisS}). Differential evolution is then used to optimize over the latent space of the autoencoder. The actual implementation for this compliance minimization problem can be found in Appendix~\ref{App:CM_implementation}.

The structural designs obtained, as well as the improvements made for $J$ during the process, are shown in~\textbf{Fig.~\ref{fig:compliance_min}}. It can be seen that the proposed procedure reliably outperforms the use of differential evolution over the search space. Further justification of the success of the proposed method can be seen in \textbf{Fig.~\ref{fig:rec_loss_comp}}, which shows the reconstruction loss and explained variance ratio of different MOR techniques (principal component analysis (PCA) and autoencoders) as a function of the latent space dimensionality $m$. This is done for both the initial random samples $\bm{X}_0$, as well as the autoencoder training set $\bm{X}_{250}$, created by optimizing each initial sample with $250$ steps of local optimization. Notice here that the locally optimized training samples require a lower latent space dimensionality $m$ to achieve the same reconstruction error as totally random samples. Furthermore, a lower $m$ is enough to achieve the same explained variance ratio, which shows that the lower reconstruction loss was not simply caused by simply moving the samples closer together in space, but rather that the training samples indeed can be found on a lower-dimensional manifold. Nonetheless, as the problems cannot be solved analytically and the search space is too large for a brute-force search, it cannot be determined if the final results found correspond to these problems' global minima.
}

\section*{Methods}  \label{sec:methods}

There are two main factors that determine if, and possibly how, an optimization problem
\begin{equation}
    \bm{x}_{\min}= \underset{\bm{x}\in X}{\text{argmin}} \; c(\bm{x}), \;\; X\subset \mathbb{R}^n
\end{equation}
can be solved with general optimization methods in a limited amount of time, namely the dimensionality $n$ of the problem and the time $t_c$ it takes to evaluate one instance of the cost function $c(\bm{x})$. Optimization problems are solved using an underlying computer architecture that, for a given period of time, imposes an upper bound on the number of possible cost function evaluations, which is approximately inversely proportional to $t_c$. As the number of cost function evaluations required to sufficiently explore the search space grows exponentially with the number of search variables $n$ (see ``curse of dimensionality''~\cite{Chen2015,Guirguis2020}, although it has to be noted that some authors seem to assume linear growth to be enough \cite{Jones1998}), using standard global optimization algorithms becomes unfeasible if the cost function evaluation time and the problem dimensionality are too high.

Since reducing $t_c$ without altering the cost function is usually not possible, optimizing previously unsolvable problems mandates for algorithms that converge with the same likelihood while requiring fewer cost function evaluations. According to the ``no free lunch theorem''~\cite{Wolpert1997,Adam2019}, this is only possible by exploiting the structure of the cost function, which will also limit the range of optimization problems to which this new algorithm can be applied. One way to exploit the problem structure is to build an algorithm that discards large regions of the search space $X$ where the global optimum $\bm{x}_{\min}$ is not likely to be found.

\begin{figure}
\centering
\tikzsetnextfilename{Figure_4}
\input{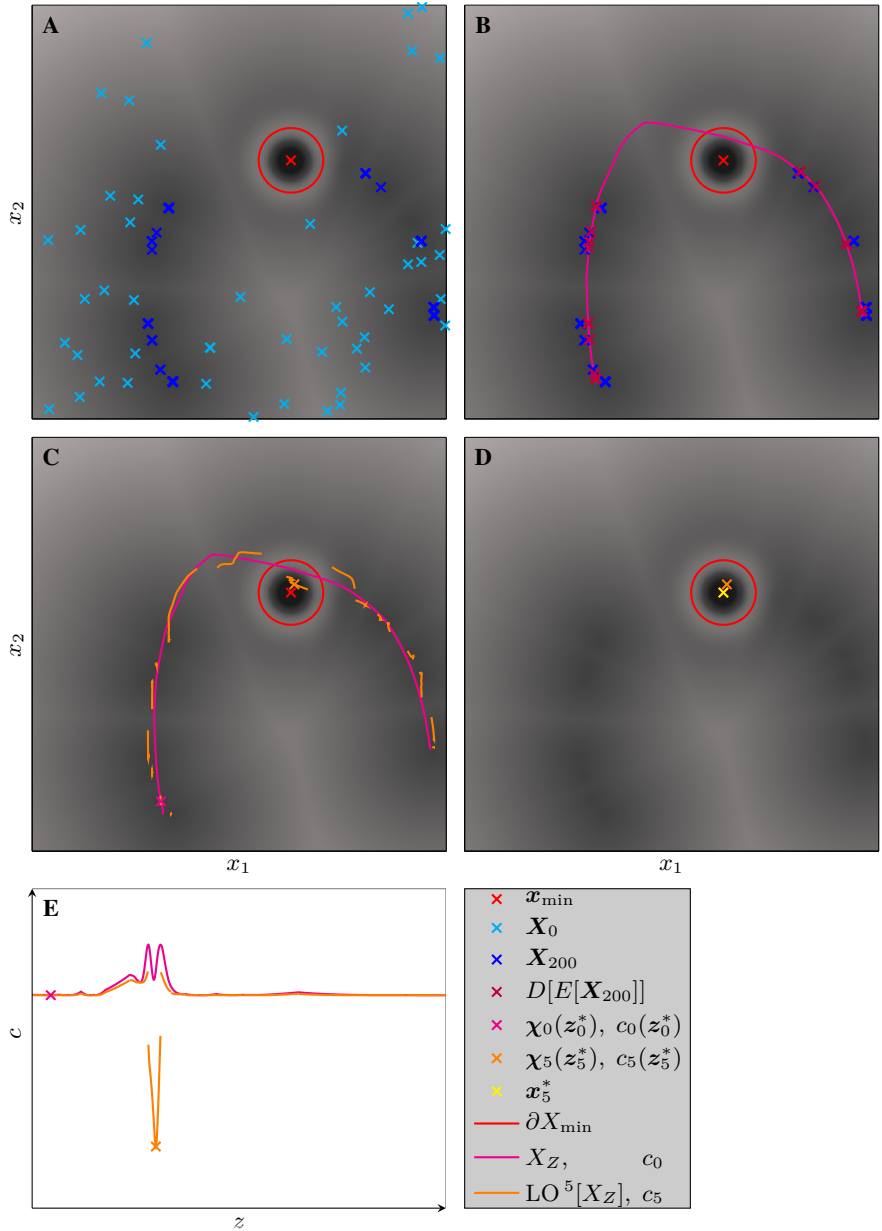}
\caption{A symbolic representation of the four steps of the proposed method for a case of $n=2$ and $m=1$. In the first step (\textbf{A}), a training set $\bm{X}_{200}$ is created by 200 iterations of gradient based local optimization, starting from random samples $\bm{X}_0$. An autoencoder is trained on this training set, leading to the decoded latent space $X_Z$ (\textbf{B}). In the third step, global optimization over latent space can be performed (\textbf{C+E}). Here, the necessity for $\mu>0$ is shown, as optimizing $c_0$ is not successful, while optimizing $c_5$ is. This can be seen in \textbf{D}, where the result of post-processing ($\bm{x}^*_5$) is nearly identical to the global minimum $\bm{x}_{\min}$.}
\label{fig:method_example}
\end{figure}

Our proposed procedure for global optimization using autoencoders uses sampled local minima to try to construct a lower-dimensional manifold that discards useless regions of the search space and that also contains the global optimum. The procedure comprises the following steps, with an schematic example of them shown in \textbf{Fig. \ref{fig:method_example}}, and a pseudocode implementation of it visible in Algorithm \ref{alg:pseudocode}.

\begin{enumerate}
    \item \textbf{Generating training samples}: A training set for the autoencoder is generated in a first step, with the goal that the training set features correspond to those of the usable part of the search space. This is done by firstly randomly sampling $N$ points $\bm{x}_i \in \mathbb{R}^n$ in the search space $X$. Local optimization is then used to advance these points in the direction of their corresponding local minima. $\bm{X}_\lambda$ denotes the set of samples after the $\lambda$th step of local optimization, \textit{i.e.}, with $\LO$ denoting the local optimization operator (which refers to any local optimization algorithm), $ \bm{X}_{\lambda}=\LO\left[\bm{X}_{\lambda-1}   \right]= \LO{}^\lambda \left[ \bm{X}_0\right].$
    The value of $\lambda$ has to be chosen carefully, since it should be large enough so that training samples $\bm{X}_{\lambda}$ are useful, and small enough to avoid unnecessary evaluations.
    
To generate these samples, we use gradient based methods like Adam~\cite{Kingma2017} or alternatively methods like SIMP~\cite{sigmund2001} or other FEM based approaches in the case of topology optimization problems. But if the cost function is not differentiable ($c(\bm{x})\not\in C^1$), zeroth order local optimization methods like the Nelder-Mead algorithm \cite{Nelder1965} can be used, although for high-dimensional problems this is likely very expensive.
It is worth noting that multiple different starting points could yield similar designs after local optimization, undoubtedly wasting computational resources. To avoid this problem, sampling techniques like deflation could be used~\cite{Papadopoulos2021}, whereby new samples are guided away from already explored regions by modifying the cost function.

 \item \textbf{Training the autoencoder}: In a second step, the samples from $\bm{X}_\lambda$ are used to create a latent space of dimensionality $m$ (see Appendix~\ref{App:AE}). While there are different related procedures (autoencoders \cite{Rumelhart1986,Kramer1991}, variational autoencoders~(VAE) \cite{Kingma2014}, generative adversarial networks (GANs)~\cite{Goodfellow2014}, adversarial autoencoders (AAEs)~\cite{Makhzani2016}, autoencoder with surrogate model network~\cite{Eismann2017}), this step produces a decoder network $D: \mathbb{R}^m \rightarrow \mathbb{R}^n$, which allows the transformation of sample $\bm{z}_i\in Z$ out of the latent space $Z\subset \mathbb{R}^m$ into the search space $X\subset \mathbb{R}^n$. Noteworthy, as illustrated earlier in  \textbf{Fig.~\ref{fig:schematic}}, decoded samples $D(\bm{z}_i) $ occupy an $m$-dimensional manifold $X_Z$ ---the decoded latent space in this work---of the $n$-dimensional search space ($X_Z=D[Z]\subset X$). Since $D\in C^1$, $X_Z$ will be a continuous domain for a continuous latent space $Z$.
 
To have the highest probability of finding the global optimum $\bm{x}_{\min}$, our hope is to construct an autoencoder whose decoded latent space $X_Z$ is as close to the global optimum $\bm{x}_{\min}$ as possible, but at least goes through the basin of the global optimum $X_{\min}$ (see Appendix \ref{App:assumption}).

{If the intrinsic dimensionality $m^*$ cannot be determined directly from the problem description, it is advisable to train multiple autoencoders with varying latent space dimensionalities $m$. One then can assume $m^*$ to be the smallest $m$ after which one cannot perceive a significant reduction in the autoencoder's reproduction loss.}

    \item \textbf{Optimization over latent space}: In this step, by optimizing over latent space, we want to find, as quickly as possible, the closest point to the global minimum. To this end, we minimize the cost function $c_\mu(\bm{z})$, \textit{i.e.},
    \begin{equation}
	\bm{z}^*_{\mu}=\underset{\bm{z}\in{Z}}{\argmin} \; c_\mu(\bm{z})\, , \;\text{with} \;\;
        c_\mu(\bm{z})=c\left(\bm{\chi}_{\mu}\left(\bm{z}\right)\right), \;\; \bm{\chi}_{\mu}\left(\bm{z}\right)= \LO{}^\mu\left( D(\bm{z}) \right), \label{eq:cost_func_simple}
    \end{equation}
	where local optimization steps ($\mu>0$) might be necessary for two reasons: Firstly, it is possible that the decoder $D$ does not generate designs $D(\bm{z})$ for random latent space samples $\bm{z}$ that fulfill all constraints of the optimization problem. The local optimization operator $\LO$ can then be used to enforce such constraints, an example of which is the mass constraint in the case of compliance minimization. And secondly, it is also possible that the decoded latent space $X_Z$ does not contain the global optimum $\bm{x}_{\min}$ but only goes through its basin $X_{\min}$. In such cases then, the best point $\bm{\chi}_{0}(\bm{z}_0^*)$ included in $X_Z$ might be far away from the global optimum $\bm{x}_{\min}$ (\textit{e.g.}, see \textbf{Fig. \ref{fig:method_example}C}). By adding $\mu$ steps of local optimization, points closer to the global optimum can be explored during the optimization over the latent space, which increases the likelihood that the optimized point $\bm{\chi}_{\mu}(\bm{z}^*_{\mu})$ reaches the global optimum at best or a point nearby at worst. {This consideration and the resulting adaption of the algorithm are to the authors' best knowledge a novel approach. }

Adding local optimization also leads to certain issues, for instance making the cost function $c_{\mu}$ discontinuous and therefore limiting the optimization algorithms that can be used. But most importantly, the computational cost associated with this third step increases approximately linearly with value of $\mu$. Depending on the time $t_c$ necessary to evaluate this cost function $c_{\mu}$, one could apply heuristic search methods like differential evolution~\cite{Storn1997} for inexpensive cost functions or Bayesian optimization~\cite{Jones1998} for more computationally involved ones. 

Consequently, an optimal balance must then be struck between speed and proximity of $\bm{\chi}_{\mu}(\bm{z}^*_{\mu})$ to the global optimum, for which $\mu$ has to be chosen carefully.

    \item \textbf{Post-processing}: A local optimum $\bm{z}^*=\text{argmin}\, c_\mu(\bm{z})$ in latent space does not necessarily correspond to a local optimum in the search space (see Appendix~\ref{App:post_explain} and \textbf{Fig. \ref{fig:method_example}D}). Consequently, the solution $\bm{\chi}_{\mu}\left(\bm{z}^*_{\mu}\right)$ can likely be improved further by advancing it further towards the final solution $\bm{x}^*$ by $\nu$ steps, resulting in
    \begin{equation}
        \bm{x}^*=\LO{}^\nu \left(\bm{\chi}_{\mu}\left(\bm{z}^*_{\mu}\right)\right). \label{eq:PP1}
    \end{equation}
    If different forms of local optimization are used for global optimization and post-processing, the following alternative is also possible, where the local optimization steps taken during the optimization over latent space are discarded:
    \begin{equation}
        \bm{x}^*=\LO{}^\nu \left(D\left(\bm{z}^*_{\mu}\right)\right). \label{eq:PP2}
    \end{equation}
\end{enumerate}

\begin{algorithm}
\caption{Pseudocode Autoencoder Enabled Gobal Optimization \label{alg:pseudocode}}

\begin{algorithmic}
\Require{Dimensionality of search space $n$ and latent space $m$; Number of training samples $N$; Number of local optimization steps $\lambda$, $\mu$, and $\nu$}
\\
\Require{Cost function $c$ with local optimization operator $\LO$; Construction, training and loss functions $AE_{\text{build}}$, $AE_{\text{fit}}$ and $\mathcal{L}$ for the autoencoder; Global optimization algorithm $\GO$}
\\
\Function{$AEGO$}{$n,m,N,\lambda,\mu,\nu, c, \LO, AE_{\text{build}}, AE_{\text{fit}},\mathcal{L} , \GO    $}
\Function{$c_\mu$}{$\bm{z},D,\mu$} \Comment{Cost function for optimization over latent space}
\State $\bm{x} \leftarrow$ \Call{$D$}{$\bm{z}$}
\For{$i \leftarrow \{1,\hdots,\mu\}$}
\State $\bm{x} \leftarrow$ \Call{$\LO$}{$\bm{x}$}
\EndFor
\State \Return \Call{$c$}{$\bm{x}$}
\EndFunction
\State \inlinecomment{\textbf{Generating training samples}}
\State $\bm{X}_\lambda \leftarrow \emptyset$
\For{$i \leftarrow  \{1,\hdots,N\}$}
\State $\bm{x} \sim U^{n}$ \Comment{Sample uniformly random initial design $\bm{x}\in \mathbb{R}^n$}
\For{$j \leftarrow  \{1,\hdots,\lambda\}$}
\State $\bm{x} \leftarrow$ \Call{$\LO$}{$\bm{x}$}
\EndFor
\State $\bm{X}_\lambda \leftarrow \left\{\bm{X}_\lambda,\bm{x} \right\}$
\EndFor
\State \inlinecomment{\textbf{Training the autoencoder}}
\State $AE\leftarrow$ \Call{$AE_{\text{build}}$}{$n,m$}  \Comment{Construct autoencoder neural network}
\State $D\leftarrow$ \Call{$AE_{\text{fit}}$}{$AE, \mathcal{L},\bm{X}_\lambda$}  \Comment{Train autoencoder $AE$ and extract trained decoder $D$}
\State \inlinecomment{\textbf{Optimization over latent space}}
\State $\bm{z}^* \leftarrow$  \Call{$\GO$}{$c_\mu,D,\mu$} \Comment{ Minimize $c_\mu$}
\State \inlinecomment{\textbf{Post-processing}}
\State $\bm{x}^* \leftarrow $  \Call{$D$}{$\bm{z}^*$}
\State $N_p \leftarrow \nu +\mu$ or $\nu$  \Comment{$N_p$ depends on choosing equation \eqref{eq:PP1} or \eqref{eq:PP2} }
\For{$i \leftarrow  \{1,\hdots,N_p\}$}      
\State $\bm{x}^* \leftarrow$ \Call{$\LO$}{$\bm{x}^*$}
\EndFor
\State \Return $\bm{x}^*$
\EndFunction
\end{algorithmic}
\end{algorithm}

\section*{Discussion}

Optimization algorithms can exploit certain properties of the underlying cost function to increase their effectiveness.
Some authors have recognized this while creating algorithms, for instance, for optimizing problems with non-deceptive gradients---\textit{i.e.}, following the average slope of the cost function leads to the global optimum~\cite{Chu2011}.
Other authors simply ignore the cost function properties and simply tweak the optimization algorithm, for example by tuning hyperparamters to achieve a faster convergence on a limited set of benchmark problems~\cite{Yuen2008,Noman2008,Du2008,Long2019,Sun2019}.
Nevertheless, because these benchmark problems are often constructed to be deceptive and/or random, the effectiveness of such procedures on real-world problems is at best questionable~\cite{Chu2011}.

Contrary to other works in the literature involving autoencoders, our proposed method of autoencoder-enabled global optimization explicitly exploits the behavior of a certain class of cost functions. For these the global optimum can be found or at least better results are obtained when compared to other well-known optimization methods. 
{For example, the proposed method can exploit the \S~\hyperref[sec:CM]{\textbf{fracture anisotropy maximization}} problem's structure because large regions of the search space are not interesting, such as any design that is mirrored along the line $y=x$ (at $45^{\circ}$), because there would be no difference in fracture resistance for different load cases, due to the different load cases being mirrored along that line as well. As a result, it can be expected that the manifold that contains the global optimum and other feasible designs, has likely a far lower intrinsic dimensionality than that of the original search space.
This assumption is supported not only by the results our proposed method can achieve, but also by \textbf{Fig. \ref{fig:rec_loss_comp}}, where it is shown that our initial assumption is likely fulfilled, since local minima have far less differentiating features compared to a set of random designs.}

The use of the proposed method, however, comes at a high cost in time and computational resources, especially when compared to gradient-based approaches (which are the standard procedure for computationally-intensive problems such as structural topology optimization~\cite{Marck2012,Zuo2017,Liu2018_2}). {For example, if one is able to intelligently choose an initial design or if the problem is unimodal, generating a design with a low cost function value would likely take less then $n_F = \num{1000}$ steps of local optimization, while for our proposed method the number of function evaluations required might be orders of magnitudes larger. Thus, the potential of our method is most promising on problems where the cost function is highly non-convex and where there are no common methods to intelligently choose an initial material distribution.} There, simple gradient-based search would likely not lead to the optimum or any comparable design~\cite{bendsoe2003,Rozvany2009,vandijk2010}. Furthermore, the proposed method becomes attractive if a wealth of feasible input data is already available, as this would drastically cut down on the time required for generating representative training samples for the autoencoder.

The proposed method however is not without flaws, as it is not always easy to determine if it can be applied at all. This results from the fact that there is no straightforward way to determine if a given cost function fulfills the underlying assumptions of the method, \textit{i.e.}, large parts of the search space are useless and can be discarded. In most cases, the only option is then logical reasoning (as for example employed for the \S~\hyperref[sec:CM]{\textbf{fracture anisotropy maximization}} problem). And even if the assumption holds, determining the optimal hyperparameters for the method remains difficult, especially in regard to the dimensionality of the latent space $m$. This is important, as decreasing $m$ will in turn reduce the likelihood of finding the global optimum, while conversely, increasing $m$ inevitably makes the optimization over latent space more time-consuming. While in this work $m$ has been chosen by repeatedly training the autoencoder with increasing values of $m$, there is further room for improvement. For example, one could test using less expensive model order reduction methods to obtain an educated estimate for the number of features in the training data that the autoencoder has to reproduce---\textit{e.g.}, non-linear PCA~\cite{Chen2017}. 

Further research could also be conducted to find possible trends for the optimal choice of other hyperparameters, like the required number of local optimization steps during the optimization over the latent space or the number of training samples, although it is possible that those will be highly dependent on the specific problem. {In addition, exploring the effects that changes of the network architecture of the autoencoder might have on the shape of the decoded latent space and its position in relation to the training samples---and their influence on the final optimized design---are other points that deserve further consideration. This might also include a renewed comparison between autoencoders and linear methods such as PCA.}
Other possible improvements are also imaginable. For example, it has been suggested that using the best designs generated with this proposed method as part of an updated sample set for a retraining of the autoencoder can lead to further improvements~\cite{Oh2019,tripp_sample-efcient_2020}. Finally, the use of faster methods to generate appropriate training samples for the neural network in the first place would also increase the viability of this approach even further.

\section*{Acknowledgments}
 The authors would like to thank the Technical University Delft for supporting this work by making available its high-performance computing cluster for their use.

\section*{Data availability statement}
The data that support the findings of this study, and the code used to generate it, have been
deposited in a { \href{https://github.com/julianschumann/aego}{public Github repository}.}

\bibliography{scibib}
\bibliographystyle{ieeetr}

\newpage

\section*{Supplementary materials}
\etocsettocstyle{\subsection*{The supplementary materials include}}{}
 \localtableofcontents
\subsection{Computational methods}
\subsubsection{Feed-forward neural networks} \label{App:FNN}
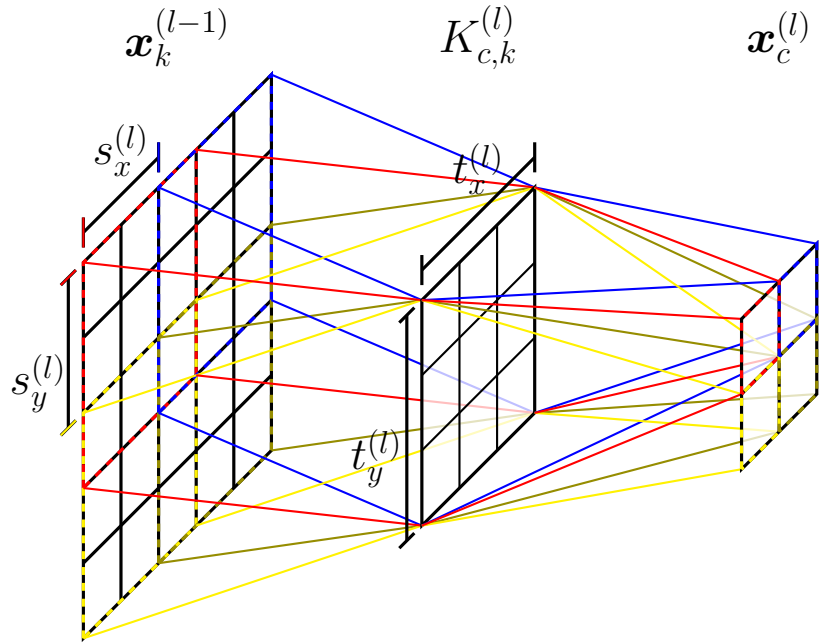
\begin{figure}[!ht]
\centering
\tikzsetnextfilename{Figure_5}
         \begin{tikzpicture}
         \filldraw[draw=black,fill=white, fill opacity=0.5,  very thick] (0,0)--(0,5)--(2.5,7.5)--(2.5,2.5)--cycle;
         \draw[black,very thick] (0.5,0.5)--(0.5,5.5);
         \draw[black,very thick] (1.0,1.0)--(1.0,6.0);
         \draw[black,very thick] (1.5,1.5)--(1.5,6.5);
         \draw[black,very thick] (2.0,2.0)--(2.0,7.0);
         \draw[black,very thick] (0,1)--(2.5,3.5);
         \draw[black,very thick] (0,2)--(2.5,4.5);
         \draw[black,very thick] (0,3)--(2.5,5.5);
         \draw[black,very thick] (0,4)--(2.5,6.5);
         \draw[red,  very thick, dashed] (0,2)--(0,5)--(1.5,6.5)--(1.5,3.5)--cycle;
         \draw[yellow,  very thick, dashed] (0,0)--(0,3)--(1.5,4.5)--(1.5,1.5)--cycle;
         \draw[blue,  very thick, dashed] (1,3)--(1,6)--(2.5,7.5)--(2.5,4.5)--cycle;
         \draw[olive,  very thick, dashed] (1,1)--(1,4)--(2.5,5.5)--(2.5,2.5)--cycle;
        
         \draw[olive, thick] (1,4)--(4.5,4.5);
         \draw[olive, thick] (1,1)--(4.5,1.5);
         \draw[olive, thick] (2.5,5.5)--(6,6);
         \draw[olive, thick] (2.5,2.5)--(6,3);
        
         \draw[blue, thick] (1,6)--(4.5,4.5);
         \draw[blue, thick] (1,3)--(4.5,1.5);
         \draw[blue, thick] (2.5,7.5)--(6,6);
         \draw[blue, thick] (2.5,4.5)--(6,3);
        
         \draw[yellow, thick] (0,3)--(4.5,4.5);
         \draw[yellow, thick] (0,0)--(4.5,1.5);
         \draw[yellow, thick] (1.5,4.5)--(6,6);
         \draw[yellow, thick] (1.5,1.5)--(6,3);
        
         \draw[red, thick] (0,5)--(4.5,4.5);
         \draw[red, thick] (0,2)--(4.5,1.5);
         \draw[red, thick] (1.5,6.5)--(6,6);
         \draw[red, thick] (1.5,3.5)--(6,3);
        
         \filldraw[draw=black,fill=white, fill opacity=0.75,  very thick] (4.5,1.5)--(4.5,4.5)--(6,6)--(6,3)--cycle;
         \draw[black, thick] (5,2)--(5,5);
         \draw[black, thick] (5.5,2.5)--(5.5,5.5);
         \draw[black, thick] (4.5,2.5)--(6,4);
         \draw[black, thick] (4.5,3.5)--(6,5);
        
         \draw[olive, thick] (4.5,4.5)--(9.25,3.75);
         \draw[olive, thick] (4.5,1.5)--(9.25,2.75);
         \draw[olive, thick] (6,6)--(9.75,4.25);
         \draw[olive, thick] (6,3)--(9.75,3.25);
        
         \draw[blue, thick] (4.5,4.5)--(9.25,4.75);
         \draw[blue, thick] (4.5,1.5)--(9.25,3.75);
         \draw[blue, thick] (6,6)--(9.75,5.25);
         \draw[blue, thick] (6,3)--(9.75,4.25);
        
         \draw[yellow, thick] (4.5,4.5)--(8.75,3.25);
         \draw[yellow, thick] (4.5,1.5)--(8.75,2.25);
         \draw[yellow, thick] (6,6)--(9.25,3.75);
         \draw[yellow, thick] (6,3)--(9.25,2.75);
        
         \draw[red, thick] (4.5,4.5)--(8.75,4.25);
         \draw[red, thick] (4.5,1.5)--(8.75,3.25);
         \draw[red, thick] (6,6)--(9.25,4.75);
         \draw[red, thick] (6,3)--(9.25,3.75);
        
         \filldraw[draw=black,fill=white, fill opacity=0.75,  very thick] (8.75,2.25)--(8.75,4.25)--(9.75,5.25)--(9.75,3.25)--cycle;
         \draw[black,very thick] (9.25,2.75)--(9.25,4.75);
         \draw[black,very thick] (8.75,3.25)--(9.75,4.25);
         \draw[red,  very thick, dashed] (8.75,3.25)--(8.75,4.25)--(9.25,4.75)--(9.25,3.75)--cycle;
         \draw[yellow,  very thick, dashed] (8.75,2.25)--(8.75,3.25)--(9.25,3.75)--(9.25,2.75)--cycle;
         \draw[blue,  very thick, dashed] (9.25,3.75)--(9.25,4.75)--(9.75,5.25)--(9.75,4.25)--cycle;
         \draw[olive,  very thick, dashed] (9.25,2.75)--(9.25,3.75)--(9.75,4.25)--(9.75,3.25)--cycle;
        
         \draw[black] (1.25,8) node[] {\Large $\bm{x}_k^{(l-1)}$};
         \draw[black] (9.25,8) node[] {\Large $\bm{x}_c^{(l)}$};
         \draw[black] (5.25,8) node[] {\Large $K_{c,k}^{(l)}$};
        
         \draw[black,very thick] (-0.2,2.8)--(-0.2,4.8);
         \draw[black,very thick] (-0.3,2.7)--(-0.1,2.9);
         \draw[yellow ,thick] (-0.3,2.7)--(-0.1,2.9);
         \draw[black,very thick] (-0.3,4.7)--(-0.1,4.9);
         \draw[red ,thick] (-0.3,4.7)--(-0.1,4.9);
        
         \draw[black,very thick] (0,5.4)--(1,6.4);
         \draw[black,very thick] (0,5.2)--(0,5.6);
         \draw[red,thick] (0,5.2)--(0,5.6);
         \draw[black,very thick] (1,6.2)--(1,6.6);
         \draw[blue,thick] (1,6.2)--(1,6.6);
        
         \draw[black] (-0.6,3.4) node[] {\Large $s_y^{(l)}$};
         \draw[black] (0.5,6.5) node[] {\Large $s_x^{(l)}$};
        
         \draw[black,very thick] (4.3,1.3)--(4.3,4.3);
         \draw[black,very thick] (4.2,1.2)--(4.4,1.4);
         \draw[black,very thick] (4.2,4.2)--(4.4,4.4);
        
         \draw[black,very thick] (4.5,4.9)--(6,6.4);
         \draw[black,very thick] (4.5,4.7)--(4.5,5.1);
         \draw[black,very thick] (6,6.2)--(6,6.6);
        
         \draw[black] (3.9,2.4) node[] {\Large $t_y^{(l)}$};
         \draw[black] (5.25,6.25) node[] {\Large $t_x^{(l)}$};
        
         \draw[white] (-1.5,-0.5) rectangle (10.5,8.5);
         \end{tikzpicture}
 
\caption{Representation of a convolutional connection between the channels $\bm{x}_k^{(l-1)}$ and $\bm{x}_c^{(l)}$, using the filter $K_{c,k}^{(l)}$, and a stride $s_y^{(l)}=s_x^{(l)}=2$, as well as a filter size $t_y^{(l)}=t_x^{(l)}=3$. In this work, $t_y^{(l)}=t_x^{(l)}=t^{(l)}$ will always be the case.\\
Here, $\bm{\vartheta}_{1,1,k}^{(l-1)}$ and $x_{1,1,c}^{(l)}$ are seen in red boundaries in $\bm{x}_k^{(l-1)}$ and $\bm{x}_c^{(l)}$ respectively. Similarly, $\bm{\vartheta}_{2,1,k}^{(l-1)}$ and $x_{2,1,c}^{(l)}$ are surrounded in yellow, $\bm{\vartheta}_{1,2,k}^{(l-1)}$ and $x_{1,2,c}^{(l)}$ in blue, and $\bm{\vartheta}_{2,2,k}^{(l-1)}$ and $x_{2,2,c}^{(l)}$ in green.}
\label{fig:fig_5}
\end{figure}

An artificial neural network is in its most basic form simply a graph of connected nodes, which in some capacity tries to model the behavior of human brains \cite{Boers1992}. In the case of a feed-forward neural network, these nodes, which are also known as neurons, each belong to different layers $\bm{y}^{(l)}$ with $l \in \{0,\hdots,L\}$. $\bm{y}^{(0)}$ is commonly called the input layer, while $\bm{y}^{(L)}$ is referred to as output layer. Layers in between are known as hidden layers. These layers are connected by often nonlinear functions $f^{(l)}_{\bm{\theta}^{(l)}}$, which are dependent on a number of parameters ${\bm{\theta}^{(l)}}$. To wit,
\begin{equation}
    \bm{y}^{(l)}=f_{\bm{\theta}^{(l)}}^{(l)}\left( \bm{y}^{(l-1)}  \right).
\end{equation} 
Consequently, a feed-forward neural network can also be seen as a number of function compositions:
\begin{equation}
    \bm{x}^{(L)}=\left(f_{\bm{\theta}^{(L)}}^{(L)}\circ \hdots \circ f_{\bm{\theta}^{(1)}}^{(1)} \right)\left( \bm{y}^{(0)}\right)
\end{equation}
Many different functions have been proposed~\cite{Oshea2015}, with most of them relying on the so called activation function $\sigma^{(l)}$, of which the following four types are used in this work:
\begin{equation}
    \begin{aligned}
        \sigma_I(x)&=x,\\
        \sigma_S(x)&={1\over{1+\exp(-x)}},\\
        \sigma_T(x)&=\tanh (x),\\
        \sigma_R(x)&=\begin{cases}x & x\geq 0,\\ 0.3 x & x<0.   \end{cases}
        \label{eq:activation_comp}
    \end{aligned}
\end{equation}
In this work, six types of connections $f^{(l)}_{\bm{\theta}^{(l)}}$ are used:
\begin{itemize}
    \item \textbf{Dense connection} One possibility to connect two layers is the dense connection ($\mathbb{O}_{\Delta}$). In this case, each element (also called neuron) $y_i^{(l-1)}$ with $i \in \{1,\hdots, \nu^{(l)}\}$ of layer $\bm{y}^{(l-1)} \in \mathbb{R}^{\nu^{(l)}}$ is connected to every neuron $y_j^{(l)}$ of the layer $\bm{y}^{(l)}$  by the weight $w_{ji}^{(l)}$ ($\bm{W}^{(l)} \in \mathbb{R}^{\nu^{(l)}\times \nu^{(l-1)}}$). Additionally, each neuron $y_j^{(l)}$ has a bias value $b_{j}^{(l)}$ ($\bm{b}^{(l)} \in \mathbb{b}^{\nu^{(l)}}$), and the activation function $\sigma^{(l)}$ is used as well:
    \begin{equation}
        \bm{y}^{(l)}=f^{(l)}_{\bm{\theta}^{(l)}}\left( \bm{y}^{(l-1)}  \right)=\sigma^{(l)}\left[ \bm{W}^{(l)} \bm{x}^{(l-1)} +\bm{b}^{(l)} \right].
    \end{equation}
    This function is therefore parameterized by $\bm{\theta}^{(l)}=\{\bm{W}^{(l)},\bm{b}^{(l)}\}$.

    \item \textbf{Convolutional connection}: A different type of connection between two layers is the convolutional one ($\mathbb{O}_{C}$). Instead of the dense layer, which is mainly used to connect two one-dimensional vectors, a convolutional function in this work connects two layers consisting of three-dimensional tensors. Each neuron $y_{ijk}^{(l)}$ is part of row $i\in \{1,\hdots, \nu_y^{(l)}\}$, column $j\in \{1,\hdots, \nu_x^{(l)}\}$, as well as channel $k\in \{1,\hdots, \kappa^{(l)}\}$, with $\bm{y}^{(l)}\in \mathbb{R}^{\nu_y^{(l)}\times \nu_x^{(l)} \times \kappa^{(l)}}$. 

    In a convolutional connection, each channel $k$ of the layer $\bm{y}^{(l-1)}$ is connected to every channel $c$ of the layer $\bm{y}^{(l)}$ by a filter $\bm{K}_{ck}^{(l)} \in \mathbb{R}^{t_y^{(l)}\times t_x^{(l)}}$. Here, $t_y^{(l)}\times t_x^{(l)}$ is the filter size. While rectangular filters are possible, it is common to use quadratic filters ($t_y^{(l)}=t_x^{(l)}=t^{(l)}$), which is done mostly in this work as well.    
    
    For every node ${y}_{ijc}^{(l)}$, a part of $\bm{x}_k^{(l-1)}$, namely $\bm{\vartheta}_{ijk}^{(l-1)}$ (chosen depending on filter size $t^{(l)}$ and stride $s^{(l)}$, see \textbf{Fig. \ref{fig:fig_5}}), has this filter applied. The results of this for all $k$ are added, and a bias $b_{ijc}^{(l)} $ is applied as well as an activation function:
    \begin{equation}
        y_{ijc}^{(l)}=\sigma^{(l)}\left(b_{c}^{(l)} + \sum\limits_{k=1}^{\kappa^{(l-1)}}      \left\langle \bm{\vartheta}_{ijk}^{(l-1)},\bm{K}_{ck}^{(l)}  \right\rangle_F  \right),
    \end{equation}
    where $\langle \, \cdot \, , \cdot\, \rangle_F$ denotes the Frobenius inner product.
    Consequently, the parameters $\bm{\theta}^{(l)}=\{\bm{K}^{(l)},\bm{b}^{(l)}\}$ define the function $f_{\bm{\theta}^{(l)}}^{(l)}$, with $\bm{K}^{(l)} \in \mathbb{R}^{\kappa^{(l)}\times \kappa^{(l-1)}\times t_y^{(l)} \times t_x^{(l)}}$.
    
    \item \textbf{Max-pooling connection}: Similar to the convolutional connection is the max pooling connection ($\mathbb{O}_{M}$). Depending on filter sizes $t_y^{(l)}$ and $t_x^{(l)}$ and stride $s^{(l)}$, a pooled area $\bm{\vartheta}_{ijk}^{(l-1)} $ is chosen. One can then determine:
    \begin{equation} \label{eq:max_pooling}
        y_{ijc}^{(l)}=\max \left( \bm{\vartheta}_{ijk}^{(l-1)}  \right).
    \end{equation}
    In this case, the function $f_{\bm{\theta}^{(l)}}^{(l)}$ has an empty parameter set $\bm{\theta}^{(l)}=\emptyset$. The activation function used here is normally the identity one $\sigma_I$, so it has not been depicted in equation~\eqref{eq:max_pooling}.
    
    \item\textbf{Deconvolutional connection}: Also used in this work is the deconvolutional connection ($\mathbb{O}_{D}$). Here, similarly to $\bm{\vartheta}_{ijk}^{(l-1)}$ and $y_{ijc}^{(l)}$ in the previous cases, $\bm{\vartheta}_{ijc}^{(l)}$ can be determined for a node $y_{ijk}^{(l-1)}$, based on filter sizes $t_y^{(l)}$ and $t_x^{(l)}$ and stride $s^{(l)}$:
    \begin{equation}
        \bm{\vartheta}_{ijc}^{(l)}=\sum\limits_{k=1}^{\kappa^{(l-1)}} \bm{K}_{ck}^{(l)}y_{ijk}^{(l-1)}.
    \end{equation}
    Finally, the channel $\bm{y}_c^{(l)}$ has to be assembled (using the assembly operator $\Aop$, which adds values from the local matrix $\bm{\vartheta}_{ijc}^{(l)}$ to the global one $\bm{y}_c^{(l)}$), with $i \in \{1,\hdots,\nu_y^{(l-1)}\}$ and $j \in \{1,\hdots,\nu_x^{(l-1)}\}$:
    \begin{equation}
        \bm{y}_c^{(l)}=\sigma^{(l)} \left[ \bm{1} b_c^{(l)} + \Aop\limits_{i,j} \bm{\vartheta}_{ijc}^{(l)} \right].
    \end{equation}
	Here, $\bm{1}$ is a matrix of the size of $\bm{y}_c^{(l)}$ with the value of $1$ at every element.
    This function is consequently parameterized by $\bm{\theta}^{(l)}=\{\bm{K}^{(l)},\bm{b}^{(l)}\}$.

A problem for deconvolutional connections is padding, as it is possible that there are elements in $\bm{y}_c^{(l)}$ for which there are no corresponding $\bm{\vartheta}_{ijc}^{(l)}$, setting them automatically to assume the value $b_c^{(l)}$. This can be avoided by using, as was done in this work, $\nu_x^{(l-1)}=s_x^{(l)}(\nu_x^{(l)}-1)+t_x^{(l)}$ and $\nu_y^{(l-1)}=s_y^{(l)}(\nu_y^{(l)}-1)+t_y^{(l)}$ for convolutional or max pooling connections, as well as $\nu_x^{(l)}=s_x^{(l)}(\nu_x^{(l-1)}-1)+t_x^{(l)}$ and $\nu_y^{(l)}=s_y^{(l)}(\nu_y^{(l-1)}-1)+t_y^{(l)}$ for deconvolutional connections.

    \item \textbf{Flattening connection}: A connection $\mathbb{O}_F$, used to flatten a layer from a three-dimensional tensor to a one-dimensional representation ($\nu_y^{(l-1)} \nu_x^{(l-1)}  \kappa^{(l-1)} =\nu^{(l)}$), is also needed:
    \begin{equation}
        \mathbb{O}_F=f^{(l)}:\mathbb{R}^{ \nu_y^{(l-1)}\times \nu_x^{(l-1)} \times \kappa^{(l-1)} } \rightarrow \mathbb{R}^{ \nu_y^{(l-1)} \nu_x^{(l-1)}  \kappa^{(l-1)} }.
    \end{equation}
    
    \item \textbf{Inverse flattening connection}: Lastly, the ability to transform a one–dimensional layer back into a three–dimensional one is also needed, for which we use $\mathbb{O}_{I}$ and which can be considered as an inversion of $\mathbb{O}_F$, with $\nu_y^{(l)} \nu_x^{(l)}  \kappa^{(l)} =\nu^{(l-1)}$ being a condition to be fullfilled:
    \begin{equation}
        \mathbb{O}_{I}(\nu_y^{(l)},\nu_x^{(l)},\kappa^{(l)})=f^{(l)}: \mathbb{R}^{\nu^{(l-1)}}\rightarrow \mathbb{R}^{ \nu_y^{(l)}\times \nu_x^{(l)} \times \kappa^{(l)}, }
    \end{equation}
    \begin{equation}
        \left(\mathbb{O}_{I}(\nu_y^{(l)},\nu_x^{(l)},\kappa^{(l)}) \circ \mathbb{O}_F\right)\left(\bm{y}^{(l)}\right)=\bm{y}^{(l)}.
    \end{equation}
\end{itemize}

\subsubsection{Autoencoder} \label{App:AE}
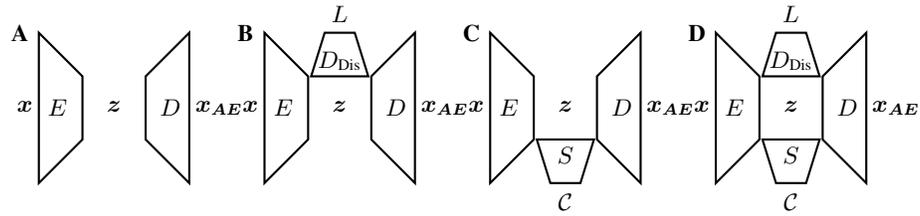
\begin{figure}[!ht]
\centering
\tikzsetnextfilename{Figure_6}
 
\begin{tikzpicture}
             \begin{scope}[xshift=9cm, yshift=0cm]
         \draw[black] (0.25,0) node[] { $E$};
         \draw[black] (1.75,0) node[] { $D$};
         \draw[black] (1,0) node[] { $\textcolor{white}{\bm{x}_{A}}\bm{z}\textcolor{white}{\bm{x}_{A}}$};
         \draw[black] (0.066,0) node[left] { $\textcolor{white}{{}_{I}}\bm{x}$};
         \draw[black] (1.933,0) node[right] { $\bm{\bm{x}_{AE}}$};
         \draw[black, thick] (0.62,0.42)--(1.38,0.42)--(1.2,1)--(0.8,1)--cycle;
         \draw[black] (1,1.25) node[] { $L$};
         \draw[black] (1,0.625) node[] { $D_{\text{Dis}}$};
         \draw[black, thick] (0.62,-0.42)--(1.38,-0.42)--(1.2,-1)--(0.8,-1)--cycle;
         \draw[black] (1,-1.25) node[] { $\mathcal{C}$};
         \draw[black] (1,-0.625) node[] { $S$};
         \node[black] at (-0.25,1) {\textbf{D}};
         \draw[black, thick] (0,-1)--(0,1)--(0.58,0.42)--(0.58,-0.42)--cycle;
         \draw[black, thick] (2,-1)--(2,1)--(1.42,0.42)--(1.42,-0.42)--cycle;
      \end{scope}
             \begin{scope}[xshift=6cm, yshift=0cm]
         \draw[black] (0.25,0) node[] { $E$};
         \draw[black] (1.75,0) node[] { $D$};
         \draw[black] (1,0) node[] { $\textcolor{white}{\bm{x}_{A}}\bm{z}\textcolor{white}{\bm{x}_{A}}$};
         \draw[black] (0.066,0) node[left] { $\textcolor{white}{{}_{I}}\bm{x}$};
         \draw[black] (1.933,0) node[right] { $\bm{\bm{x}_{AE}}$};
         \draw[black] (1,1.25) node[] { \textcolor{white}{$L$}};
         \draw[black, thick] (0.62,-0.42)--(1.38,-0.42)--(1.2,-1)--(0.8,-1)--cycle;
         \draw[black] (1,-1.25) node[] { $\mathcal{C}$};
         \draw[black] (1,-0.625) node[] { $S$};
         \node[black] at (-0.25,1) {\textbf{C}};
         \draw[black, thick] (0,-1)--(0,1)--(0.58,0.42)--(0.58,-0.42)--cycle;
         \draw[black, thick] (2,-1)--(2,1)--(1.42,0.42)--(1.42,-0.42)--cycle;
      \end{scope}
             \begin{scope}[xshift=3cm, yshift=0cm]
         \draw[black] (0.25,0) node[] { $E$};
         \draw[black] (1.75,0) node[] { $D$};
         \draw[black] (1,0) node[] { $\textcolor{white}{\bm{x}_{A}}\bm{z}\textcolor{white}{\bm{x}_{A}}$};
         \draw[black] (0.066,0) node[left] { $\textcolor{white}{{}_{I}}\bm{x}$};
         \draw[black] (1.933,0) node[right] { $\bm{\bm{x}_{AE}}$};
         \draw[black, thick] (0.62,0.42)--(1.38,0.42)--(1.2,1)--(0.8,1)--cycle;
         \draw[black] (1,1.25) node[] { $L$};
         \draw[black] (1,0.625) node[] { $D_{\text{Dis}}$};
         \draw[black] (1,-0.625) node[] { \textcolor{white}{$S$}};
         \node[black] at (-0.25,1) {\textbf{B}};
         \draw[black, thick] (0,-1)--(0,1)--(0.58,0.42)--(0.58,-0.42)--cycle;
         \draw[black, thick] (2,-1)--(2,1)--(1.42,0.42)--(1.42,-0.42)--cycle;
      \end{scope}
             \begin{scope}[xshift=0cm, yshift=0cm]
         \draw[black] (0.25,0) node[] { $E$};
         \draw[black] (1.75,0) node[] { $D$};
         \draw[black] (1,0) node[] { $\textcolor{white}{\bm{x}_{A}}\bm{z}\textcolor{white}{\bm{x}_{A}}$};
         \draw[black] (0.066,0) node[left] { $\textcolor{white}{{}_{I}}\bm{x}$};
         \draw[black] (1.933,0) node[right] { $\bm{\bm{x}_{AE}}$};
         \draw[black] (1,1.25) node[] { \textcolor{white}{$L$}};
         \draw[black] (1,-1.25) node[] { \textcolor{white}{$\mathcal{C}$}};
         \node[black] at (-0.25,1) {\textbf{A}};
         \draw[black, thick] (0,-1)--(0,1)--(0.58,0.42)--(0.58,-0.42)--cycle;
         \draw[black, thick] (2,-1)--(2,1)--(1.42,0.42)--(1.42,-0.42)--cycle;
      \end{scope}
\draw[white] (-0.5,-1.5) rectangle (11.5,1.5);
\end{tikzpicture}
 
\caption{A depiction of different autoencoder networks with encoder $E$ and decoder $D$. In \textbf{A}, a simple autoencoder is seen. In \textbf{B}, through the addition of a discriminator network $D_{\text{Dis}}$ an adversarial autoencoder is seen. On the other hand, in \textbf{C}, one added a surrogate network $S$ instead. In \textbf{D}, the final possible combination is depicted, an adversarial autoencoder with surrogate network.}
\label{fig:fig_6}
\end{figure}

An autoencoder is a special type of feed-forward neural network (see Appendix~\ref{App:FNN}), first proposed by Rumelhart \textit{et al.}~\cite{Rumelhart1986} and improved by Kramer~\cite{Kramer1991}. An autoencoder, in its most basic form, is distinguished by two characteristic properties: 
\begin{itemize}
    \item Firstly, the input layer $\bm{y}^{(0)}$ and the output layer $\bm{y}^{(L)}$  have the same dimensionality. In this work, the input layer will be denoted as $\bm{x}=\bm{y}^{(0)}$, while the output layer will be $\bm{x}_{\text{AE}}=\bm{y}^{(L)}$. The goal of the autoencoder is to be trained in such a way to minimize the difference between input and output for some training data.
    \item An autoencoder has at least one hidden layer ($L>1$) with fewer nodes than those of input and output layers. The hidden layer $l_B$ with the lowest number of nodes, which is commonly known as the bottleneck layer, is usually  named $\bm{z}=\bm{y}^{(l_B)}\in \mathbb{R}^m$.
\end{itemize}
It is possible to split the autoencoder network into two separate networks, the encoder $E$ and the decoder $D$ (see \textbf{Fig. \ref{fig:fig_6}A}):
\begin{equation}
    \begin{aligned}
    E_{\bm{\theta}_E}=&\left(f_{\bm{\theta}^{(l_B)}}^{(l_B)}\circ \hdots \circ f_{\bm{\theta}^{(1)}}^{(1)} \right), \;\;\;\;\;\, \bm{\theta}_E=\left\{\bm{\theta}^{(1)},\hdots,\bm{\theta}^{(l_B)}   \right\}, \\
    D_{\bm{\theta}_D}=&\left(f_{\bm{\theta}^{(L)}}^{(L)}\circ \hdots \circ f_{\bm{\theta}^{(l_B+1)}}^{(l_B+1)} \right),\;\; \bm{\theta}_D=\left\{\bm{\theta}^{(l_B+1)},\hdots,\bm{\theta}^{(L)}   \right\}. \\
    \end{aligned}
\end{equation}
In this work, the domain of input and output layers is referred to as the search space $X$ ($\bm{x},\bm{x}_{\text{AE}}\in X$), while the domain of $\bm{z}$ is the latent space $Z$ ($\bm{z}\in Z$).
The following can then be assumed:
\begin{equation}
    \bm{x}_{\text{AE}}=D_{\bm{\theta}_D}(\bm{z})=D_{\bm{\theta}_D}(E_{\bm{\theta}_E}(\bm{x}))
\end{equation}
The corresponding reconstruction loss function $\mathcal{L}_{R,i}$ for a single sample $\bm{x}_i$ is then the mean squared error, where $\bm{W}_i$ is the corresponding weight matrix:
\begin{equation}
    \mathcal{L}^{E,D}_{\text{R},i}=\langle\bm{W}_i,\left(\bm{x}_i-D(E(\bm{x}_i))\right)^2 \rangle_F. \label{eq:loss_Rec}
\end{equation}
The $E,D$ in the superscript of the loss function indicate which network parameters are optimized depending on this loss, which in this case are the encoder $E$ and decoder $D$ networks.
Depending on the number of hidden layers, autoencoders can be discriminated into deep ($L>2$) and shallow ($L=2$) autoencoders~\cite{Charte2018}.
\label{App:DisS}
Although using only encoder $E$ and decoder $D$ networks is possible, expanding the network might be advantageous, with different possibilities and their benefits being explored in the following paragraphs~\cite{Eismann2017,Wang2020,Kudyshev2020}. Four different network architectures have been proposed (see \textbf{Fig. \ref{fig:fig_6}}):
\begin{itemize}
    \item When using an additional discriminator network $D_{\text{Dis}}$, the expanded architecture is known as an adversarial autoencoder~\cite{Makhanzi2015}. The main purpose of its use is to enforce a certain distribution of the encoded training samples in latent space, which would lead to feasible designs being produced over the whole latent space~\cite{Kudyshev2020}. Additionally, using a discriminator might also lead to a more even distribution of local minima and maxima of the function $c_\mu(\bm{z})$ (see Equation~\eqref{eq:cost_func_simple}), which would be advantageous for  a successful optimization over latent space. While adversarial autoencoders are similar in that purpose to generative adversarial networks~\cite{Goodfellow2014}, they do not suffer from mode collapse~\cite{Mondal2020}, meaning that they do not randomly loose features of the training data, and allow for easier training due to the lower-dimensional input of the discriminator network~\cite{Kudyshev2020}. To achieve this, the discriminator maps a latent space representation $\bm{z}$ onto the likelihood $L=D_{\text{Dis}}(\bm{z}) \in [0,1]$ that it has been generated according to the desired probability distribution $\mathcal{Z}_D$, which might be normal or uniform distribution (although more complex ones, especially for labeled data, can be used~\cite{Makhanzi2015}). To achieve this, the discriminator is trained to differentiate between two different sources for latent space samples, namely random samples ($\bm{z}_i \sim \mathcal{Z}_D$) and encoded training set samples ($E(\bm{x}_i)$ with $\bm{x}_i \in \bm{X}_{\lambda}$).
    The training optimizes the parameters $\bm{\theta}_{D_{\text{Dis}}}$ of the discriminator by means of the following loss function:
    \begin{equation}
        \mathcal{L}^{D_{\text{Dis}}}_{\text{D1},i}=- {1\over{2}} \left(\ln \left(D_{\text{Dis}}(E(\bm{x}_i))\right)+\ln \left(1-D_{\text{Dis}}(\bm{z}_i)\right)\right),\label{eq:loss_Dis1}
    \end{equation}
    This forces the discriminator to put out $D_{\text{Dis}}(E(\bm{x}_i))\rightarrow 0^+$ for samples originating from the training set, while samples generated according to the desired distribution are forced towards $D_{\text{Dis}}(\bm{z}_i)\rightarrow 1^-$.

    The encoder meanwhile is trained to fool the discriminator by minimizing the loss function
    \begin{equation}
        \mathcal{L}^{E}_{\text{D2},i}=- \ln \left(1-D_{\text{Dis}}(E(\bm{x}_i))\right),\label{eq:loss_Dis2}
    \end{equation}
    so that $D_{\text{Dis}}(E(\bm{x}_i))$ tends towards $L=1$.
    These two loss functions are then each added to the overall loss function of the network for each batch with a probability of $P_{\text{Dis}}$, so that none, one or both of the loss functions can be used to update the network parameters in each batch.
 The corresponding network architectures that use $D_{\text{Dis}}$ are shown schematically in \textbf{Figs. \ref{fig:fig_6}B} and \textbf{Figs. \ref{fig:fig_6}D}.
    \item One can also add a surrogate network $S$, which maps a latent space vector $\bm{z}$ onto the cost function approximation $\mathcal{C}=S(\bm{z}) \in [0,1]$. This could enforce a improved separation between better and worse samples in latent space, possibly reducing the number of local minima for $c(D(\bm{z}))$, which would allow for a faster optimization~\cite{Eismann2017,Wang2020}. The surrogate network is then trained to reproduce the scaled cost function value $c(\bm{x})$:
    \begin{equation}
    \begin{aligned}
        c_{n,i}&={1\over{10}} + {8\over{10}} {c(\bm{x}_i)-c_{\min} \over{c_{\max}-c_{\min}}},\\
        c_{\min}&= \underset{\bm{x}_i\in \bm{X}_{\lambda}}{\min} c(\bm{x}_i),\\
        c_{\max}&= \underset{\bm{x}_i\in \bm{X}_{\lambda}}{\max} c(\bm{x}_i).
    \end{aligned}
    \end{equation}
    The following loss function is then used to train the encoder $E$ and the surrogate model $S$ to that end:
    \begin{equation}
        \mathcal{L}^{E,S}_{\text{surr},i}=\beta_{S} \left(c_{n,i}-S(E(\bm{x}_i))\right)^2.  \label{eq:loss_Surr}
    \end{equation}
    This loss function is added to the overall loss function of the whole network. Network architectures that use $S$ are shown schematically in \textbf{Fig. \ref{fig:fig_6}C} and \textbf{Fig. \ref{fig:fig_6}D}. 
\end{itemize}

\label{App:NNTraining}To train the neural networks in this work, backpropagation and gradient based optimization is used. This makes the performance of the trained networks dependent on the initial parameters set in the networks. Bias vectors were initialized as zero, while weight tensors $\bm{W}^{(l)}$ and filters $\bm{K}^{(l)}$ are set using Xavier initialization~\cite{Glorot2010}. After initializing the network parameters $\bm{\theta}_{0}$, batch gradient descent is used to optimize the network parameters over multiple epochs. Adam~\cite{Kingma2017} is used to update the network parameters after every batch from, using a learning rate of $\alpha=0.001$, as well as decay rates $\beta_1=0.9$ and $\beta_2=0.999$.

\subsubsection{Pretraining} \label{App:Pretraining}
For autoencoder networks, there is always the possibility of using pretraining for the encoder $E$ and decoder $D$.This has been shown to improve the autoencoder performance, although there is conflicting evidence~\cite{Bengio2007,Zhou2015}.

For pretraining to be possible, the encoder $E$ and decoder $D$ need layers with the same dimensionality. This would then allow one to split up both into $n_{\text{pre}}$ smaller networks $E_i$ and $D_i$, where the input layer of $E_i$ and the output layer of $D_i$ would have the same dimension, as well as the output layer of $E_i$ and the input layer of $D_i$:
\begin{equation}
\begin{aligned}
    E&=\left( E_{n_{\text{pre}}} \circ \hdots \circ E_{1} \right)\\
    D&=\left( D_1 \circ \hdots \circ D_{n_{\text{pre}}} \right)
\end{aligned}
\end{equation}
One could then train these smaller networks, beginning with $E_1$ and $D_1$ and then going inwards from there, by minimizing either the loss function
\begin{equation}
    \mathcal{L}^{E_j,D_j}_{\text{pre}1,i}=\rho\left(\left(\bm{x}_i-\left(D_1\circ \hdots \circ  D_j \circ    E_j \circ \hdots\circ E_{1} \right)(\bm{x}_i)\right)^2\right)   \label{eq:pretraining_loss_2}
\end{equation}
or
\begin{equation}
    \mathcal{L}^{E_j,D_j}_{\text{pre}2,i}=\rho\left(\left(\left(E_{j-1} \circ \hdots\circ E_{1} \right)(\bm{x}_i)-\left(D_j \circ    E_j \circ \hdots\circ E_{1} \right)(\bm{x}_i)\right)^2 \right). \label{eq:pretraining_loss}
\end{equation}
While the first loss function overall leads to smaller losses, this is achieved by a longer calculation time. Therefore, in this work, $\mathcal{L}_{\text{pre}2}$ is used for pretraining.



\subsection{Implementation of numerical simulations}
\subsubsection{Minimizing $c_1$,  $c_2$, $c_3$, and $c_4$ from \S~\hyperref[sec:BM_test]{\textbf{Benchmark functions}}} 
\label{App:POC1} \label{App:POC2}
{
The function $c_1$ is designed in multiple steps to test the viability of the proposed method.
\begin{enumerate}
    \item In an $M=5$ dimensional space ($\mathcal{Z}=[-1,1]^{5}$), $\num{1000}$ points $\bm{\zeta}_i\in \mathcal{Z}$ are generated randomly according to an uniform distribution. These points $\bm{\zeta}=\{\bm{\zeta}_1,\hdots, \bm{\zeta}_{\num{2500}}\}$ are then saved,  with $\bm{\zeta}_1={1\over{10}} \cdot \bm{1}$ ($\bm{1}$ is a vector where every element is equal to 1).
    \item A mapping $T_D: \mathbb{R}^M \rightarrow \mathbb{R}^n$ is created, which allows to find $\bm{x}_i \in \mathbb{R}^n$ from $\bm{\zeta}_i$ with $n=\num{100}$:
    \begin{equation}\begin{aligned}
    \left[{x}_{1,i}\right]_j=& \begin{cases}  \left[\zeta_{i}\right]_j & j\leq M  \\ 0 & j> M  \end{cases}      ,\\
    \bm{x}_{2,i}=&  \tanh \left[ \bm{W} \bm{x}_{1,i}  \right],\\
    \left[x_{3,i}\right]_j=& {9\over{10}}{\left[x_{2,i}\right]_j\over{ \underset{i}{\max} \; \vert \left[x_{2,i}\right]_j \vert    }} ,\\
    \left[ x_i \right]_j=&  \left[x_{3,i}\right]_j+  {1\over{10}} \left(  1- {\left[x_{3,i}\right]_j}^2   \right)  .
    \end{aligned}
    \end{equation}
    This transformation $T_D$---where $\bm{W} \in \mathbb{R}^{n \times n}$ is a randomly generated rotation matrix---then allows the generation of $\bm{X}=T_D[\bm{\zeta}]=\{\bm{x}_1,\hdots, \bm{x}_{\num{2500}}\}$.
    
    \item A first function $f_1$ is then built as
    \begin{equation}
        f_1(\bm{x})= \underset{i\in \{2, \hdots, 1000\}}{\min} \, \vert\vert \bm{x}-\bm{x}_i \vert\vert^2,
    \end{equation}
    which has multiple local minima, with $f_1(\bm{x}_i)=0$ $\forall i \in \{2, \hdots, 1000\} $.
\item A second function $f_2$ is then defined, with $c_{0,i} \sim U\left(1,2\right)$:
\begin{equation}
	f_2(\bm{x}) =     {  \sum\limits_{i=2}^{1000}    1 + {c_{0,i}\over{\vert\vert \bm{x}-\bm{x}_i \vert\vert^2}} \over{ \sum\limits_{i=2}^{1000}  {1\over{\vert\vert \bm{x}-\bm{x}_i \vert\vert^2}}  }}
\end{equation}
The point $f_2(\bm{x}_i)= c_{0,i}$ $\forall i \in \{2, \hdots, 1000\}$ are the local minima here, the same as for $f_1$.
    \item A final function $f_3$
    \begin{equation}
        f_3(\bm{x})= \min \left\{ 1, {1\over{R^2}} \vert\vert \bm{x}-\bm{x}_1 \vert\vert^2 \right\}
    \end{equation} 
    is constructed, where $R={1\over{2}}$ limits the support of the gradient $\nabla f_3$. This function has one global optimum at $\bm{x}_{1}$. Local optimization should lead to the global optimum if initial guesses are within a distance smaller than $R$ from it (this is the basin $X_{\min}$). Local optimization with initial guesses further away will most likely not converge to the global optimum. For  $\num{5000}$ training samples, the likelihood $P_g$ to randomly sample the global optimum is then (using the first order Taylor polynomial)
\begin{equation} \begin{aligned}
P_g&=1-(1-p)^{5000}\approx 5000 p\approx 7.5 \times  10^{-97} \\    \text{for} \;\; p&\approx {\vert X_{\min} \vert \over{\vert X \vert}}={ {\pi^{n\over{2}} \over{ \Gamma\left({n\over{2}}+1 \right)}} R^{n} \over{ 2^{n}}}={ {\pi^{50} \over{ \Gamma\left(51 \right)}}{1\over{2}}^{100} \over{ 2^{100}}}\approx 1.5 \times 10^{-100}
\end{aligned}\end{equation}
    \item The final cost function $c$, with the global minimum at $\bm{x}_{1}$, is then given by
    \begin{equation}
        c_1(\bm{x})=\left(5 f_1(\bm{x})+f_2(\bm{x})\right) f_3(\bm{x}).
    \end{equation}
This cost function than has a global optimum with $c(\bm{x}_1)=0$, while all other local minima have at least a cost function value of $c_1\geq 1$, as long as $R\leq \underset{ i\in \{2, \hdots, 1000\} }{\min} \, \vert\vert \bm{x}_1-\bm{x}_i \vert\vert$.
\end{enumerate}}
Additionally, three benchmark functions from a paper by Abualigah \textit{et al.}~\cite{abualigah2021} were also considered:
\begin{equation}
    \begin{aligned}
    c_2(\bm{x})=& \sum\limits_{i=1}^n \left(-x_i \sin\left( \sqrt{\vert x_i\vert}\right) \right)+418.9829 n,\\
    c_3(\bm{x})=&{\pi\over{n}}\left( 10 \sin\left(\pi y(x_1)\right)^2 + \sum\limits_{i=1}^{n-1} \left(y(x_i)-1\right)^2 \left(1+ 10 \sin\left(\pi y(x_{i+1})\right)^2    +u (\bm{x})   \right)  \right),\\
    &y(x)={x+5\over{4}},\;\; u(\bm{x})=\sum\limits_{i=1}^n  100 \max\left\{0,\vert x_i\vert-10\right\}^4,\\
    c_4(\bm{x})=& 1+{1\over{4000}}\sum\limits_{i=1}^n\left( x_i^2\right)-\prod\limits_{i=1}^n\left( \cos\left({x_i\over{\sqrt{i}}}\right)\right).
    \end{aligned}
\end{equation}

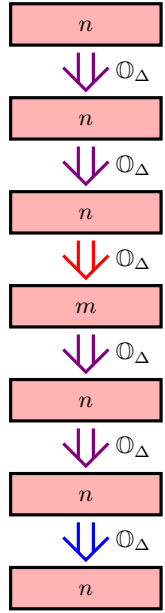
\begin{figure}[!ht]
\centering
\tikzsetnextfilename{Figure_7}
          \begin{tikzpicture}
          
             \draw[white] (-1.75,0.5) rectangle (3.75,-8.5);
            
             \begin{scope}[xshift=0cm, yshift=0cm]
        
             \def\Ty{{0,0,0,0,0,0}}
             \def\Tx{{0,0,0,0,0,0}}
             \def\St{{0,0,0,0,0,0}}
             \def\Nx{{0,0,0,0,0,0}}
             \def\Ka{{0,0,0,0,0,0}}
             \filldraw[draw=black,fill=red!30!white,  very thick] (0,0)--(0,-0.55)--(2,-0.55)--(2,0)--cycle;
             \foreach \typ/\col [count=\i from 0] in {F/violet, F/violet, F/red, F/violet, F/violet, F/blue}
             {
                 \pgfmathsetmacro{\h}{-1.25*\i-0.55}
                 \pgfmathsetmacro{\ty}{\Ty[\i]}
                 \pgfmathsetmacro{\tx}{\Tx[\i]}
                 \pgfmathsetmacro{\st}{\St[\i]}
                 \pgfmathsetmacro{\nx}{\Nx[\i]}
                 \pgfmathsetmacro{\ka}{\Ka[\i]}
                 \pgfmathsetmacro{\ha}{\h -0.1}
                 \pgfmathsetmacro{\hb}{\h -0.3}
                 \pgfmathsetmacro{\hc}{\h -0.5}
                 \pgfmathsetmacro{\hd}{\h -0.6}
                 \pgfmathsetmacro{\he}{\h -0.7}
                 \pgfmathsetmacro{\hf}{\h -0.975}
                 \pgfmathsetmacro{\hg}{\h -1.25}
                 \pgfmathsetmacro{\hh}{\h -0.35}
                 \draw[\col,very thick] (0.9,\ha)--(0.9,\hc);
                 \draw[\col,very thick] (1.1,\ha)--(1.1,\hc);
                 \draw[\col,very thick] (0.7,\hb)--(1,\hd)--(1.3,\hb);
                 \draw[black] (1.25,\hh) node[right] { $\mathbb{O}_{\Delta}$};
                 \filldraw[draw=black,fill=red!30!white,  very thick] (0,\he)--(2,\he)--(2,\hg)--(0,\hg)--cycle;
                
             }
             \draw[black] (1,-0.3) node[] { $n$};
             \draw[black] (1,-1.55) node[] { $n$};
             \draw[black] (1,-2.8) node[] { $n$};
             \draw[black] (1,-4.05) node[] { $m$};
             \draw[black] (1,-5.3) node[] { $n$};
             \draw[black] (1,-6.55) node[] { $n$};
             \draw[black] (1,-7.8) node[] { $n$};
         
         \end{scope}
         \end{tikzpicture}
 
\caption{Neural network used in section \ref{App:POC1} (see \textbf{Fig. \ref{fig:fig_10}} for symbol explanation). Red arrows indicate the activation function $\sigma_S$, blue ones $\sigma_T$, and orange ones $\sigma_R \circ \sigma_T$ (see equation \eqref{eq:activation_comp}). $n$ and $m$ are respectively the dimensionality of the search space and the latent space.}
\label{fig:fig_7}
\end{figure}

For each of the four benchmark function, the search space dimension is chosen to be $n=100$, but with different domains ($X_1=[-1,1]^n$, $X_2=[-500,500]^n$, $X_3=[-50,50]^n$, $X_4=[-500,500]^n$), while for all these functions the minimum cost function value is $c_i=0$. This global minimum is then to be found using two different approaches:
\begin{itemize}
    \item Firstly, the proposed method from \S~\hyperref[sec:methods]{\textbf{Methods}} is used, with the four steps being implemented as follows:
    \begin{enumerate}
        \item $\num{5000}$ random samples are created by sampling the search space $X_i$ uniformly. From these points, the training set $\bm{X}_{100}$ is created using Adam with differing parameters (see \textbf{Tab. \ref{tab:tab_2}}).
        \item A neural network (see \textbf{Fig. \ref{fig:fig_7}}) is then trained with mean squared error loss function for 100 epochs, with 20 batches each. The latent space dimensionality $m$ is varied, with $m\in \left\{ 2, \hdots, 7 \right\}$.
      
        \item The optimization over latent space is then performed, using the cost function $c_{\mu}$ (see Equation \eqref{eq:cost_func_simple}) with $\mu=5$. We used differential evolution $\text{DE}_{5 m,1000,0.6,0.95}$ for the optimization, and Adam local optimization steps $\LO$ (see \textbf{Tab. \ref{tab:tab_6}}).
Here,
\begin{equation}
\text{DE}_{\gamma,G,F,\chi_0} \label{eq:DE}
\end{equation}
is the notation for setting the hyperparameters of differential evolution~\cite{Storn1997}. These are the population size $\gamma$, the number of generations $G$, the multiplication factor $F$ for the generation of offspring and the probability $\chi_0$ for the crossover operator. In this work, $c_\mu$ is always optimized over the domain $[0,1]^m$.
        \item Finally, during post-processing, Adam was used for $\nu=\num{1000}$ iterations with parameters from \textbf{Tab. \ref{tab:tab_3}}
        \end{enumerate}
    \item {Alternatively, an optimization over the whole latent space is performed using differential evolution. Here, $\text{DE}_{100,10000,0.6,0.95}$ is used.}
\end{itemize}

It has to be noted that in \textbf{Fig. \ref{fig:benchmark_results}} (as well as later in \textbf{Fig. \ref{fig:compliance_min}}), the number of function evaluations $n_{F}$ is a discrete number, but due to the high number of available data points, and to better distinguish between different steps of the process with different types of lines, continuous lines have been used.

\subsubsection{Tailoring fracture resistance} \label{App:CM_implementation}

\begin{figure}[!ht]
\centering
\tikzsetnextfilename{Figure_9}
\begin{tikzpicture}
             
             \begin{scope}[xshift=0cm, yshift=0cm]
             \node[black] at (-0.5,0.5) {\textbf{A}};
             \def\Ty{{4,4,4,0,0,0,0,0,0,4,4,4}}
             \def\Tx{{4,4,4,0,0,0,0,0,0,4,4,4}}
             \def\St{{1,2,2,0,0,0,0,0,0,2,2,1}}
             \def\Ny{{30,14,6,144,84,50,84,144,6,14,30,27}}
             \def\Nx{{30,14,6,0,0,0,0,0,6,14,30,27}}
             \def\Ka{{1,4,4,0,0,0,0,0,4,4,1,1}}
             \filldraw[draw=black,fill=red!50!white,  very thick] (0.1,-0.4)--(0.2,0.1)--(2.1,0.1)--(2.1,-0.4)--cycle;
             \filldraw[draw=black,fill=red!30!white,  very thick] (0,0)--(0,-0.5)--(2,-0.5)--(2,0)--cycle;
             \draw[black] (1,-0.3) node[] { $27,27,1$};
             \foreach \typ/\col [count=\i from 0] in {D/olive, C/olive, C/olive, F/black, F/olive, F/red, F/olive, F/olive, F/black, D/olive,  D/olive, C/blue}
             {
                 \pgfmathsetmacro{\h}{-1.25*\i-0.55}
                 \pgfmathsetmacro{\ty}{\Ty[\i]}
                 \pgfmathsetmacro{\tx}{\Tx[\i]}
                 \pgfmathsetmacro{\st}{\St[\i]}
                 \pgfmathsetmacro{\ny}{\Ny[\i]}
                 \pgfmathsetmacro{\nx}{\Nx[\i]}
                 \pgfmathsetmacro{\ka}{\Ka[\i]}
                 \expandafter \ifnum \nx = 0
	                 \pgfmathsetmacro{\ha}{\h -0.1}
	                 \pgfmathsetmacro{\hb}{\h -0.3}
	                 \pgfmathsetmacro{\hc}{\h -0.5}
	                 \pgfmathsetmacro{\hd}{\h -0.6}
	                 \pgfmathsetmacro{\he}{\h -0.7}
	                 \pgfmathsetmacro{\hf}{\h -0.975}
	                 \pgfmathsetmacro{\hg}{\h -1.25}
	                 \pgfmathsetmacro{\hh}{\h -0.35}
                     \draw[\col,very thick] (0.9,\ha)--(0.9,\hc);
                     \draw[\col,very thick] (1.1,\ha)--(1.1,\hc);
                     \draw[\col,very thick] (0.7,\hb)--(1,\hd)--(1.3,\hb);
                    
                     \pgfmathsetmacro{\nxm}{\Nx[\i-1]}
                     \expandafter \ifnum \nxm = 0
                         \draw[black] (1.25,\hh) node[right] { $\mathbb{O}_{\Delta}$};
                     \else
                         \draw[black] (1.25,\hh) node[right] { $\mathbb{O}_{F}$};
                     \fi
                     \filldraw[draw=black,fill=red!30!white,  very thick] (0,\he)--(2,\he)--(2,\hg)--(0,\hg)--cycle;
                     \draw[black] (1,\hf) node[] { $\ny$};
                 \else
	                 \pgfmathsetmacro{\ha}{\h -0.1}
	                 \pgfmathsetmacro{\hb}{\h -0.2}
	                 \pgfmathsetmacro{\hc}{\h -0.4}
	                 \pgfmathsetmacro{\hd}{\h -0.5}
                     \pgfmathsetmacro{\he}{\h -0.6}
                     \pgfmathsetmacro{\hf}{\h -0.7}
                     \pgfmathsetmacro{\hg}{\h -0.975}
                     \pgfmathsetmacro{\hh}{\h -1.15}
                     \pgfmathsetmacro{\hi}{\h -1.25}
                     \pgfmathsetmacro{\hj}{\h -0.3}
                     \draw[\col,very thick] (0.9,\ha)--(0.9,\hc);
                     \draw[\col,very thick] (1.1,\ha)--(1.1,\hc);
                     \draw[\col,very thick] (0.7,\hb)--(1,\hd)--(1.3,\hb);
                    
                     \expandafter \ifnum \i = 0 
                         \draw[black] (1.25,\hj) node[right] { $\mathbb{O}_{\typ}(\ty,\tx,\st)$};
                     \else
                         \pgfmathsetmacro{\nxm}{\Nx[\i-1]}
                         \expandafter \ifnum \nxm = 0
                             \draw[black] (1.25,\hj) node[right] { $\mathbb{O}_{I}$};
                         \else
                             \draw[black] (1.25,\hj) node[right] { $\mathbb{O}_{\typ}(\ty,\tx,\st)$};
                         \fi
                     \fi
                    
                     \filldraw[draw=black,fill=red!50!white,  very thick] (0.1,\he)--(2.1,\he)--(2.1,\hh)--(0.1,\hh)--cycle;
                     \filldraw[draw=black,fill=red!30!white,  very thick] (0,\hf)--(2,\hf)--(2,\hi)--(0,\hi)--cycle;
                     \draw[black] (1,\hg) node[] { $\ny,\nx,\ka$};
                 \fi
             }
        \end{scope}
             
             \begin{scope}[xshift=5cm, yshift=-7.5cm] 
             \node[black] at (-0.5,0.5) {\textbf{B}};
             \def\Ty{{0,0,0,0,0,0}}
             \def\Tx{{0,0,0,0,0,0}}
             \def\St{{0,0,0,0,0,0}}
             \def\Ny{{50,50,50,50,50,1}}
             \def\Nx{{0,0,0,0,0,0}}
             \def\Ka{{0,0,0,0,0,0}}
             \filldraw[draw=black,fill=red!30!white,  very thick] (0,0)--(0,-0.55)--(2,-0.55)--(2,0)--cycle;
             \draw[black] (1,-0.3) node[] { $25$};
             \foreach \typ/\col [count=\i from 0] in {F/blue, F/blue, F/blue, F/blue, F/blue, F/red}
             {
                 \pgfmathsetmacro{\h}{-1.25*\i-0.55}
                 \pgfmathsetmacro{\ty}{\Ty[\i]}
                 \pgfmathsetmacro{\tx}{\Tx[\i]}
                 \pgfmathsetmacro{\st}{\St[\i]}
                 \pgfmathsetmacro{\ny}{\Ny[\i]}
                 \pgfmathsetmacro{\nx}{\Nx[\i]}
                 \pgfmathsetmacro{\ka}{\Ka[\i]}
                 \pgfmathsetmacro{\ha}{\h -0.1}
                 \pgfmathsetmacro{\hb}{\h -0.3}
                 \pgfmathsetmacro{\hc}{\h -0.5}
                 \pgfmathsetmacro{\hd}{\h -0.6}
                 \pgfmathsetmacro{\he}{\h -0.7}
                 \pgfmathsetmacro{\hf}{\h -0.975}
                 \pgfmathsetmacro{\hg}{\h -1.25}
                 \pgfmathsetmacro{\hh}{\h -0.35}
                 \draw[\col,very thick] (0.9,\ha)--(0.9,\hc);
                 \draw[\col,very thick] (1.1,\ha)--(1.1,\hc);
                 \draw[\col,very thick] (0.7,\hb)--(1,\hd)--(1.3,\hb);
                 \draw[black] (1.25,\hh) node[right] { $\mathbb{O}_{\Delta}$};
                 \filldraw[draw=black,fill=red!30!white,  very thick] (0,\he)--(2,\he)--(2,\hg)--(0,\hg)--cycle;
                 \draw[black] (1,\hf) node[] { $\ny$};
                
             }
         \end{scope} 
         \end{tikzpicture}
 
\caption{Neural networks used in section \ref{App:CM_implementation} (see \textbf{Fig. \ref{fig:fig_10}} for symbol explanation). Red arrows indicate the activation function $\sigma_S$, blue ones $\sigma_T$, olive ones $\sigma_R$, and black ones $\sigma_I$ (see equation \eqref{eq:activation_comp}). In \textbf{A},  the encoder and decoder (with $m=50$) can be seen, while in \textbf{B}, the discriminator and surrogate model network (again for $m=50$)---which have the same architecture---are depicted.}
\label{fig:fig_9}
\end{figure}
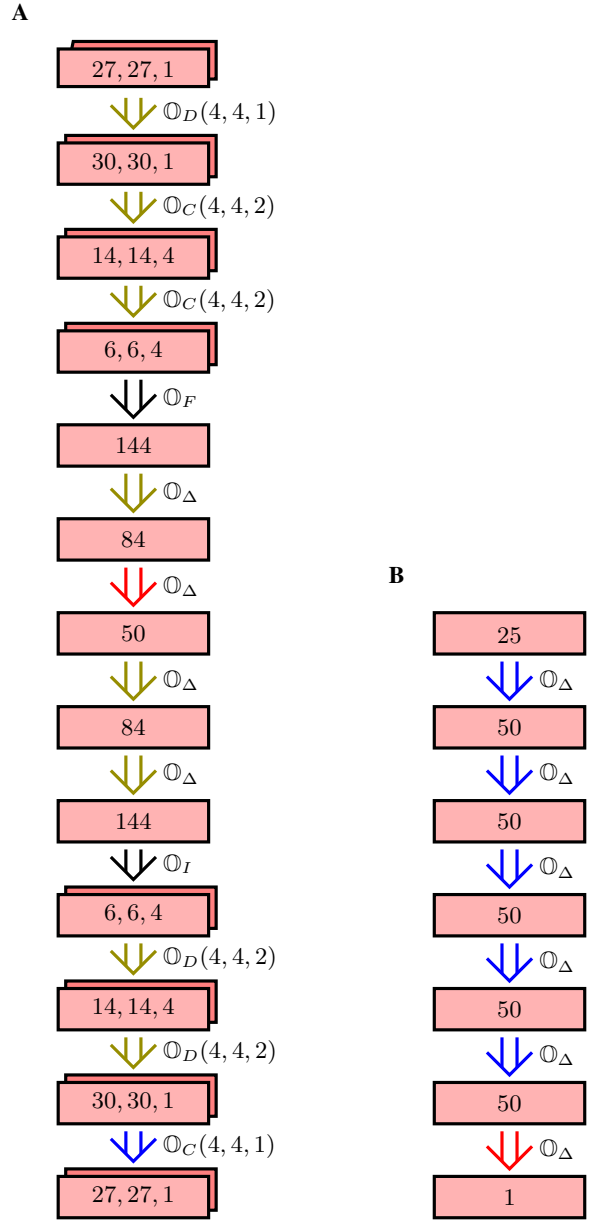

{The exact implementation of the four steps of the proposed method (see \S~\hyperref[sec:methods]{\textbf{Methods}}) for tailoring the fracture resistance of a unit cell (see \S~\hyperref[sec:CM]{\textbf{fracture anisotropy maximization}}) is now presented in detail. 
Alternatively to the proposed method---to create a baseline---differential evolution with $DE_{300,300,0.6,0.9}$ is also run over the search space.

\paragraph{Training data set} \label{App:CM_training_set}
The local optimization method $\LO$ is used to generate $\num{6000}$ training samples for the autoencoder, with $\lambda = 250$, resulting in $\bm{X}_{250}$. For the original samples $\bm{X}_0 = \left\{\bm{x}_{0,1}, \hdots \bm{x}_{0,5000}\right\}$, they are created using 
\begin{equation}
	\left[\bm{x}_{0,1}\right]_{k}= U(-1,1) \left[\bm{x}_B\right]_{k} \,, \label{eq:random_mean}
\end{equation}
with the method for creating the boundary values $\bm{x}_B$, as well as the exact implementation of $\LO$ discussed by  Souto \textit{et al.}~\cite{souto_edible_2022}. 

\paragraph{Training the autoencoder} \label{App:CM_AE_training}
After generating the training set, its samples are used to train an adversarial autoencoder (the discriminator network enforces a uniform distribution over the latent space, with $P_{\text{Dis}}=3/10$) with a surrogate model network $S$ (with the weight $\beta_S=1/4$) (for more information see \textbf{Fig. \ref{fig:fig_6}D} and Appendix~\ref{App:DisS}). The specific network architectures used can be seen in \textbf{Figs. \ref{fig:fig_9}A} and \textbf{Figs. \ref{fig:fig_9}B}, with a layer architecture mainly chosen based on the results of prior simulations. Here, the main nonlinearity used in the autoencoder is the LeakyRelu function $\sigma_R$, while sigmoid function $\sigma_S$ and \textit{tangens hyperbolicus} $\sigma_T$ are used to enforce the constraints of latent and search space on the outputs of encoder and decoder respectively (see equation \eqref{eq:activation_comp}).
 During training, all elements of all samples are treated equally, with $w_j = 1$. The latent space dimensionality of $m=50$ is chosen due to the visible kink in the reproduction loss curve at this point in \textbf{Fig.~\ref{fig:rec_loss_comp}}. 
Furthermore, 100 epochs of pretraining (see Appendix~\ref{App:Pretraining}) are used for the encoder network $E$ and decoder network $D$, with 60 batches. During the pretraining, each part of the network is split into $n_{\text{pre}}=2$ parts, with the outer part consisting of the convolutional layers of the network, and the inner part consisting of fully connected layers (see \textbf{Fig. \ref{fig:fig_9}A}).
After that, 200 epochs with 60 batches each are used to train the whole network architecture (see Appendix~\ref{App:NNTraining}).

\paragraph{Optimization over latent space} \label{App:CM_optimizaion}
When optimizing over latent space, the cost function 
\begin{equation}
    c_{\mu}(\bm{z})=\left(J_i \circ
    \LO{} ^\mu \circ D
    \right) \left(\bm{z}\right).
\end{equation}
was optimized with $\mu=100$ using differential evolution $\text{DE}_{100,250,0.6,0.9}$ (see Equation \eqref{eq:DE}).
Here, the inclusion of a significant number of steps of local optimization helps to enforce the mass constraint, which is not always enforced by decoded designs $D(\bm{z})$.
 
\paragraph{Post-processing}
From the optimum $\bm{z}^{*}_{\mu}$, the final design $\bm{x}^{*}$ is obtained as
\begin{equation}
    \bm{x}^{*}= \left(\LO {}^{\mu + \nu} \circ D
    \right) \left(\bm{z}^{*}_{\mu}\right).
\end{equation}
Therefore, the post-processing method given by Equation~\eqref{eq:PP1} with $\nu = 500$ is used.}


\subsection{Supplementary text}
\subsubsection{Intrinsic dimensionality of an optimization problem} 											\label{App:assumption}
We try to construct an autoencoder with a decoder $D_{\bm{\theta}}: \mathbb{R}^m \rightarrow \mathbb{R}^n$, which is parameterized with the weights and biases $\bm{\theta}\in \bm{\Theta}$, which allows to construct a decoded latent space $X_Z(\bm{\theta})$, optimizing over which enables us to find the global optimum $\bm{x}_{\min}$. For this to be possible, $\bm{\theta} \in \bm{\Theta}^{*}$ is necessary, with
\begin{equation} \label{eq:assum1}
\bm{\Theta}^{*}=\left\{\bm{\theta} : X_Z(\bm{\theta}) \cap X_{\min} \neq \emptyset \right\} \subset \bm{\Theta},
\end{equation}
where $X_{\min}$ is the basin of $\bm{x}_{\min}$, \textit{i.e.},
\begin{equation}
X_{\min}=\left\{\bm{x} : \underset{\nu\rightarrow \infty}{\lim} \left(\LO{}^\nu (\bm{x})\right)=\bm{x}_{\min}   \right\} \subset X.
\end{equation}
Here it has to be mentioned that theoretically one could also use $\bm{\Theta}^{*}=\left\{\bm{\theta} : \bm{x}_{\min} \in X_Z(\bm{\theta}) \right\} $, but due to the inclusion of local optimization steps during the optimization over latent space and the use of post-processing (see  \S~\hyperref[sec:methods]{\textbf{Methods}}), the softer criterion given by \eqref{eq:assum1} can be used.

Before training the autoencoder, an initial $\bm{\theta}$ has to be chosen according to a prior probability distribution $p_0(\bm{\theta})$. During training then, these parameters are changed according to the autoencoder loss function $\mathcal{L}$ and the training samples $\bm{X} =\left\{\bm{x}_i \sim p_{\bm{X}}(\bm{x}): i=\{1,\hdots N\} \right\}$, resulting in the posterior $p(\bm{\theta})$ (which depends on $p_0(\bm{\theta})$, $p_{\bm{X}}$ and $\mathcal{L}$ as well as the training algorithm used). This posterior would also theoretically allow the calculation of $p_{X_Z}(\bm{x})$
\begin{equation}
p_{X_Z}(\bm{x}) = P(\bm{x} \in X_Z)  = \int\limits_{\{\bm{\theta}: \bm{x} \in   X_Z(\bm{\theta}) \}} p(\bm{\theta})  \dd\bm{\theta}
\end{equation}

One can then calculate the probability of a successful optimization run $P^*$ in two ways:
\begin{equation} \begin{aligned}
	P^*&=P(\bm{\theta}\in \bm{\Theta}^*)=\int\limits_{\bm{\Theta}^*} p(\bm{\theta})  \dd\bm{\theta}\\
		&=P(X_Z \cap X_{\min} \neq \emptyset) =\int\limits_{X_{\min}} p_{X_Z}(\bm{x}) \dd\bm{x}
\end{aligned}\end{equation} 
This probability $P^{*}(m)$ depends on the latent space dimensionality $m$, and it can be assumed that it will grow with it, as expanding $m$ will likely grow $\bm{\Theta}^{*}$ (Decreasing $m$ can be just seen as fixing certain elements of $\bm{\theta}$, so the same parameter space can be used).
Consequently, for a given set minimum probability $P_{\min}$ where we see the likelihood of successfully finding the global minimum as sufficiently high, one can find the intrinsic dimensionality $m^{*}$ of the optimization problem considered as
\begin{equation}
m^{*}=\min\left\{ m: P^{*}(m)>P_{\min}  \right\},
\end{equation}
It has to be noted that besides $m$, $P^*$ will also depend on other variables, namely  $p_0(\bm{\theta})$, $p_{\bm{X}}$, $\mathcal{L}$, and the autoencoder training algorithm. Our assumption that now underlies our proposed method is that $m^{*}\ll n$. 

It has to be noted that, while $P_{\min}$ is inherently subjective, an objective lower bound can be given when comparing the proposed method to another global optimization algorithm. For the proposed method, this would be $P_{\min}^{-1} t_{\text{PM}}$ (see Appendix \ref{App:power}), where $t_{\text{PM}}$ is the time required for one run of our proposed method, while $t_{\text{GO}}$ is the average time needed for the compared algorithm to successfully find the global optimum. Then 
\begin{equation}
	P_{\min}>{t_{\text{PM}}\over{t_{\text{GO}}}}.
\end{equation}
{Finally, it is also important to recognize that $P^{*}$ is only an upper bound to the real-world probability of finding the global optimum, as it would require the brute force optimization of $c_{\mu \rightarrow \infty}$ (see Equation~\eqref{eq:cost_func_simple}) to guarantee the finding of $\bm{x}_{\min}$ ($\bm{x}_{\min}= D_{\bm{\theta}} (\bm{z}^{*}_{\mu \rightarrow \infty})$ for $X_Z \cap X_{\min} \neq \emptyset$). This is impossible with limited computational resources due to two reasons:
\begin{itemize} 
\item We cannot optimize $c_{\mu}$ with $\mu\rightarrow \infty$, but have to use instead a finite number of local optimization steps $\mu$. This means that there is the possibility that $D_{\bm{\theta}} (\bm{z}^{*}_{\mu})\not\in X_{\min}$, especially if there are other local minima with cost function values only slightly larger than the global minimum or if the distance from $X_Z$ to $\bm{x}_{\min}$ is larger than the distance from $X_Z$ to other local minima.

\item One cannot use brute force search over a continuous domain $Z$, and therefore has to use an heuristic optimization algorithm instead. As this algorithm can become trapped in a local minimum of $c_{\mu}$, the probability of finding $\bm{z}^{*}_{\mu}$ is hence likely somewhat lower than $1$.
\end{itemize}

In specific cases, different and more direct definitions of the intrinsic dimensionality $m^*$ could be used as well, such as in the case of functions that change along a linear subspace~\cite{wang_bayesian_2016}.}

\subsubsection{On the average number of tries until success} \label{App:power}
The average number $a$ of tries necessary to get an event with a single instance likelihood of $p$ has to be calculated. We can define the probability $P(a)$ that an event will not happen $a-1$ consecutive times and then happen on the $a$th try as
\begin{equation}
P(a)=p(1-p)^{a-1}.
\end{equation}
The average number of tries $a^{*}$ is then:
\begin{equation}
\begin{aligned}
a^{*}&=\sum\limits_{a=1}^\infty a P(a)\\
&=\sum\limits_{a=1}^\infty a p(1-p)^{a-1}\\
&={p\over{1-p}}\sum\limits_{a=1}^\infty a(1-p)^a\\
&={p\over{1-p}} {1-p\over{((1-p)-1)^2}}\\
&={1\over{p}}.
\end{aligned}
\end{equation}
For the step from third to fourth line, see the $\mathrm{Li}_{-1}$ polylogarithm.

\subsubsection{Intrinsic dimensionality of $c_1$ from  \S~\hyperref[sec:BM_test]{\textbf{Benchmark functions}}}\label{App:proof_c1}
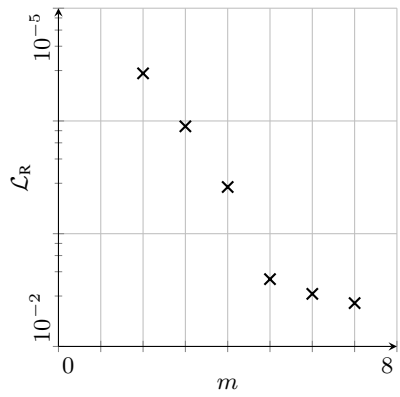
\begin{figure}[!ht]
\centering
\tikzsetnextfilename{Figure_13}
        \begin{tikzpicture}
        \begin{semilogyaxis}[
            at = {(1cm,1cm)},
            width=4.5cm,
            height=4.5cm,
	    	 scale only axis=true,
                 axis lines = left,
            enlargelimits=false,
            xlabel = ,
            ylabel = ,
            xmin=0,
            xmax=8,
            xtick={0,1,2,3,4,5,6,7,8},
                 xticklabels={\empty},
		 scaled x ticks = false,
            ymin=1e-5,
            ymax=1e-2,
            ytick={1e-5,1e-4,1e-3,1e-2},
                 yticklabels={\empty},
                 minor ytick={2.8e-5,4.6e-5,6.4e-5,8.2e-5,   2.8e-4,4.6e-4,6.4e-4,8.2e-4,     2.8e-3,4.6e-3,6.4e-3,8.2e-3},
                 yminorgrids=false,
                 xmajorgrids=true,
		 ymajorgrids=true,
                 legend style={fill=white, fill opacity=0, draw opacity=1, text opacity=1,draw=none}, 
                 legend pos=north west,
                 legend cell align={left}
            ]

        \addplot+[
            only marks,
            mark=x,
            black,
            mark options={fill=black, fill opacity=0.8,solid},
            mark size=3 pt, thick]
        table[x=m, y=e]
        {Figure_2/f_1error.txt};
        
        \end{semilogyaxis}
        \node[black,right] at (0.9,0.75) {\noexpand\noexpand\noexpand\footnotesize{$0$}};
	 \node[black,left] at (5.6,.75) {\noexpand\noexpand\noexpand\footnotesize{$8$}};
	 \node[black] at (3.25,0.5) {$m$};

	 \node[black,right,rotate=90] at (0.75,0.9) {\noexpand\noexpand\noexpand\footnotesize{$10^{-2}$}};
	 \node[black,left, rotate=90] at (0.75,5.6) {\noexpand\noexpand\noexpand\footnotesize{$10^{-5}$}};
	 \node[black, rotate=90] at (0.5,3.25) {$\mathcal{L}_{\text{R}}$};
        
        \draw[white] (0.1,0.1) rectangle(5.6,5.6);
       
        \end{tikzpicture}
\caption{The reconstruction loss $\mathcal{L}_{\text{R}}$ after training the autoencoders with varying latent space dimensionality $m$ for the optimization of the cost function $c_1$.}
\label{fig:fig_13}
\end{figure}

{
In this problem, the cost function is designed so that it has it local minima aligned on a $5$-dimensional domain inside the higher-dimensional search space $X$ (see Appendix \ref{App:POC1}). The existence of the lower dimensional manifold is confirmed in \textbf{Fig. \ref{fig:fig_13}}, where it can be seen that the reconstruction error $\mathcal{L}_{\text{R}}$ of the autoencoder after learning on the training samples decreases markedly slower for $m>5$, being close to machine precision for $m\geq 5$.

Due to the method of generating training samples, the distribution $p_{\bm{X}}$ defined in Appendix \ref{App:assumption} will be high at and around these local minima. If the autoencoder's latent space dimensionality $m$ now matches---or surpasses---the dimensionality $M=5$ of the manifold containing the local minima and the autoencoder training is successful, then one should see (with $\varsigma$ as scaling factor), that
\begin{equation}
  p_{X_Z}\approx \varsigma p_{\bm{X}}.
\end{equation}
Based on this, one can then calculate the likelihood $P^*$ of including the decoded latent space in the basin of the global optimum:
\begin{equation}
P^*=P(X_Z \cap X_{\min} \neq \emptyset) =\int\limits_{X_{\min}} p_{X_Z}(\bm{x}) \, \dd{\bm{x}} \approx \int\limits_{X_{\min}} \varsigma  p_{\bm{X}}(\bm{x}) \, \dd{\bm{x}}.
\end{equation}
As $p_{\bm{X}}$ is high around all local minima and therefore also around the global optimum, it can be assumed that $P^*$ will be high enough for $m\geq 5$ to successfully find the global optimum. 

Meanwhile, if $m<5$, the autoencoder cannot capture all training samples fully---as seen in the far higher reconstruction error for $m<5$ in \textbf{Fig. \ref{fig:fig_13}}. That means that $p_{X_Z}$ will now be far lower than compared to the case of $m \geq 5$ in the areas where $p_{\bm{X}}$ is high, as depending on the parameter initialization, the autoencoder will switch between closely reproducing only random subsets of the found training samples. Consequently, $p_{X_Z}$ will also be far lower over $X_{\min}$, resulting also in a very low $P^*$, which explains the failure to find the global optimum for $m<5$ with the proposed method.

It can therefore be explained why the intrinsic dimensionality $m^*$ of $c_1$ is identical to the dimensionality $M$ of the manifold holding the local minima, with $m^*=M=5$.
}

\subsubsection{Intrinsic dimensionality of $c_3$ and $c_4$ from  \S~\hyperref[sec:BM_test]{\textbf{Benchmark functions}}} 				\label{App:proof_eq_mean}
In \S~\hyperref[sec:BM_test]{\textbf{Benchmark functions}} we observed the tendency of autoencoders encoding evenly spread samples $\bm{X}$ from an $n$-dimensional manifold onto an $m$-dimensional latent space $Z$ to include the mean of the samples $\overline{\bm{x}}$ in the decoded latent space $X_Z$, which allowed the successful optimization of $c_3$ and $c_4$. This can be explained by the fact that the mean has the lowest reconstruction error of any point when encoding a random cluster into a single point, which incentivizes the autoencoder to include this point in its decoded latent space $X_Z$:
\begin{equation}
    \overline{\bm{x}}=\underset{\bm{x}\in X}{\text{argmin}}\; \sum\limits_i \left\Vert\bm{x}_i -\bm{x}\right\Vert^2
\end{equation}
This can be proven by firstly considering a one-dimensional problem:
\begin{equation}
\begin{aligned}
    \sum\limits_i (x_i-x)^2&=\sum\limits_i \left(x_i^2 -2x_i x +x^2\right)\\
    &= \sum\limits_i \left(x_i^2\right) - 2x \sum\limits_i \left(x_i\right) + n x^2\\
    &=\sum\limits_i \left(x_i^2\right) - n\overline{x}^2+n\overline{x}^2            -2nx\overline{x} +nx^2\\
    &= \sum\limits_i \left(x_i^2\right) - n\overline{x}^2+ n(x-\overline{x})^2\\
    &= n(x-\overline{x})^2+c
\end{aligned}
\end{equation}
This can the be extrapolated in a higher dimensional domain:
\begin{equation}\begin{aligned}
    \underset{{x}}{\text{argmin}} \sum\limits_i ({x}_i-{x})^2 &= \underset{{x}}{\text{argmin}} \; n(x-\overline{x})^2+c=\overline{x}\\
\underset{\bm{x}}{\text{argmin}} \sum\limits_i \vert\vert\bm{x}_i-\bm{x}\vert\vert^2&=\underset{\bm{x}}{\text{argmin}} \sum\limits_{j=1}^n  \sum\limits_i \left\Vert \left[x_i\right]_j-\left[x\right]_j \right\Vert^2  = \overline{\bm{x}}.
\end{aligned}\end{equation}

{As the local minima are now approximately uniformly distributed, it should be possible to rotate the decoded latent shape, in whatever form, around the mean $\overline{\bm{x}}$ without changing the reconstruction loss. Consequentially, every rotation angle should be equally feasible, which supports the assumption
\begin{equation}
p_{X_Z}(\bm{x})\approx {\varsigma_1\over{\Vert \overline{\bm{x}}-\bm{x}\Vert^{n-m}}}, \label{eq:random_train_XZ}
\end{equation}
using the probabilistic terms from Appendix~\ref{App:assumption}, where $\varsigma_1$ is a scaling constant. From this, it is clear that the likelihood for successful optimization }
\begin{equation}
	P^{*}=P(X_Z \cap X_{\min} \neq \emptyset) =\int\limits_{X_{\min}} p_{X_Z}(\bm{x}) \, \dd{\bm{x}} \approx {\varsigma_2\over{\Vert \overline{\bm{x}}-\bm{x}\Vert^{n-m}}}.
\end{equation}
is high for a global minimum close to the center $\overline{\bm{x}}$, and consequently $m^{*}\ll n$ in this cases, with $m^{*}$ being the lower the larger $X_{\min}$ is in relation to the search space $X$. In contrast, $P^{*}$ is rapidly decreasing when the global optimum moves away from the center $\overline{\bm{x}}$. Consequently, in most cases, where $\overline{\bm{x}}\not\approx \bm{x}_{\min}$ and $p_{\bm{X}}$ is approximately uniform (the means regular distribution of local minima in $X$), one can find $m^{*}=n$. 

{This is supported by the results of $c_2$, where the global optimum lies close to the boundary $\partial X$ of the search space $X$ and where consequentially no global optimum could be found with the proposed method.}

\subsubsection{On the need for post-processing} \label{App:post_explain}
It can be assumed that the result $\bm{z}^{*}$ of the third step of the proposed method will be a local optimum of $c_\mu$ in latent space. But the decoded sample $\bm{\chi}_{\mu}\left(\bm{z}^{*} \right)$ would not necessarily correspond to a local minimum of $c$. Mathematically,
\begin{equation}\begin{aligned}
\left.{\partial c_\mu\left(\bm{z}\right)\over{\partial \bm{z}}}\right\vert_{\bm{z}=\bm{z}^{*}} =\left.{\partial c\left( \bm{\chi}_{\mu}\left(\bm{z}\right)\right)\over{\partial \bm{z}}}\right\vert_{\bm{z}=\bm{z}^{*}}=&\left. {\partial c\left( \bm{x}\right)\over{\partial \bm{x}}}\right\vert_{\bm{x}=\bm{\chi}_{\mu}\left(\bm{z}^{*} \right)} \left.{\partial \bm{\chi}_{\mu}\left(\bm{z}\right)\over{\partial \bm{z}}}\right\vert_{\bm{z}=\bm{z}^{*}}=\bm{0} \\
\not\Rightarrow \;&\left. {\partial c\left( \bm{x}\right)\over{\partial \bm{x}}}\right\vert_{\bm{x}=\bm{\chi}_{\mu}\left(\bm{z}^{*} \right)} =\bm{0}.
\label{eq:proof_pp}
\end{aligned} \end{equation}
For easier notation,
\begin{equation}
\left. {\partial c\left( \bm{x}\right)\over{\partial \bm{x}}}\right\vert_{\bm{x}=\bm{\chi}_{\mu}\left(\bm{z}^{*} \right)}=\bm{b}\in \mathbb{R}^{1\times n}
\end{equation}
and
\begin{equation}
\left.{\partial \bm{\chi}_{\mu}\left(\bm{z}\right)\over{\partial \bm{z}}}\right\vert_{\bm{z}=\bm{z}^{*}}=\bm{A}\in\mathbb{R}^{n \times m}
\end{equation}
are defined. One then can proof equation \eqref{eq:proof_pp} by recognizing $\bm{b}\bm{A}=\bm{0}$ as a homogeneous system of linear equations---\textit{i. e.}, it is consistent and the trivial solution $\bm{b}=\bm{0}$ will always work. But for $m<n$, $\bm{b}\bm{A}=\bm{0}$ is an underdetermined system of linear equations, which means that an infinite number of other solutions exist for this problem besides $\bm{b}=\bm{0}$. Consequently, the claim $\bm{b}\bm{A}=\bm{0}\Rightarrow \bm{b}=\bm{0}$ is indeed incorrect.



\clearpage

\subsection{Figures}

\clearpage 

\clearpage

\clearpage

\clearpage
\begin{figure}[!ht]
\centering
\tikzsetnextfilename{Figure_10}
\begin{tikzpicture}
             \draw[white] (-0.75,5) rectangle (4.75,-5);
             \begin{scope}[xshift=0cm, yshift=0cm]
             \filldraw[draw=black,fill=red!50!white,  very thick] (0.2,3.2)--(0.2,4.7)--(3.2,4.7)--(3.2,3.2)--cycle;
             \filldraw[draw=black,fill=red!30!white,  very thick] (0,3)--(0,4.5)--(3,4.5)--(3,3)--cycle;
             \draw[black] (1.5,3.75) node[] { $\nu_y^{(l-1)},\nu_x^{(l-1)},\kappa^{(l-1)}$};
             \draw[black] (3.3,3.85) node[right] { \LARGE$\bm{x}^{(l-1)}$};
            
             \filldraw[draw=black,fill=red!50!white,  very thick] (0.2,0.2)--(0.2,1.7)--(3.2,1.7)--(3.2,0.2)--cycle;
             \filldraw[draw=black,fill=red!30!white,  very thick] (0,0)--(0,1.5)--(3,1.5)--(3,0)--cycle;
             \draw[black] (1.5,0.75) node[] { $\nu_y^{(l)},\nu_x^{(l)},\kappa^{(l)}$};
             \draw[black] (3.3,0.85) node[right] { \LARGE$\bm{x}^{(l)}$};

             \draw[black,very thick] (1.4,2.8)--(1.4,1.9);
             \draw[black,very thick] (1.6,2.8)--(1.6,1.9);
             \draw[black,very thick] (1.2,2.1)--(1.5,1.8);
             \draw[black,very thick] (1.8,2.1)--(1.5,1.8);
             \draw[black] (2,2.35) node[right] { $\mathbb{O}(t_y^{(l)},t_x^{(l)},s^{(l)})$};
             \node[black] at (-0.5,2.35) {\textbf{A}};
        
          \end{scope}
             \begin{scope}[xshift=0cm, yshift=-4.5cm]   
             \filldraw[draw=black,fill=red!30!white,  very thick] (0,2.5)--(0,3.5)--(3,3.5)--(3,2.5)--cycle;
             \draw[black] (1.5,3) node[] { $\nu^{(l-1)}$};
             \draw[black] (3.3,3) node[right] { \LARGE $\bm{x}^{(l-1)}$};
            
             \filldraw[draw=black,fill=red!30!white,  very thick] (0,0)--(0,1)--(3,1)--(3,0)--cycle;
             \draw[black] (1.5,0.5) node[] { $\nu^{(l)}$};
             \draw[black] (3.3,0.5) node[right] { \LARGE$\bm{x}^{(l)}$};

             \draw[black,very thick] (1.4,1.3)--(1.4,2.3);
             \draw[black,very thick] (1.6,1.3)--(1.6,2.3);
             \draw[black,very thick] (1.2,1.5)--(1.5,1.2);
             \draw[black,very thick] (1.8,1.5)--(1.5,1.2);
             \draw[black] (2,1.75) node[right] { $\mathbb{O}_{\Delta}$};
             \node[black] at (-0.5,1.75) {\textbf{B}};
        
        \end{scope} 
         \end{tikzpicture}
 
\caption{Symbolic representation of different connections from layer $\bm{x}^{(l-1)}$ to $\bm{x}^{(l)}$. In \textbf{A}, the depiction is of either convolutional ($\mathbb{O}=\mathbb{O}_C$), max pooling ($\mathbb{O}=\mathbb{O}_M$), or deconvolutional ($\mathbb{O}=\mathbb{O}_D$) connections. In \textbf{B}, the depiction is of a dense connection, marked with $\mathbb{O}_{\Delta}$. The activation function used will be marked by the color of the arrow connecting the different layers.}
\label{fig:fig_10}
\end{figure}
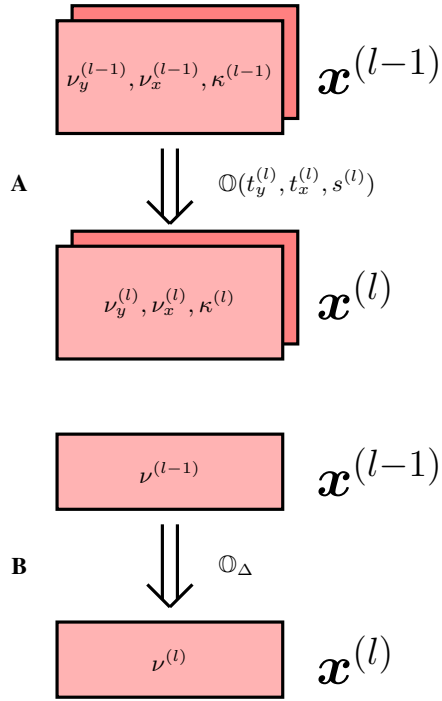

\clearpage


\clearpage
\subsection{Tables}

\noindent {\bf Tab. 2.} Parameters of Adam used for local optimization during the generation of the training set for the optimization of test functions $c_2$, $c_3$, and $c_4$.
\begin{table}[!ht]
\refstepcounter{table}
\centering

\begin{tabularx}{5.5cm}{X|YYY}
         \toprule $c_i$  & $\alpha$ & $\beta_1$ & $\beta_2$  \\ \midrule
         $c_1$ & $0.02$ & $0.5$ & $0.75$ \\ 
         $c_2$ & $20$ & $0.9$ & $0.999$ \\ 
         $c_3$ & $3$ & $0.5$ & $0.75$ \\ 
         $c_4$ & $30$ & $0.9$ & $0.999$ \\ \bottomrule
\end{tabularx}

\label{tab:tab_2}
\end{table}

\noindent {\bf Tab. 3.} Parameters of Adam used for local optimization during the optimization over latent space for the optimization of test functions $c_2$, $c_3$, and $c_4$.
\begin{table}[!ht]
\refstepcounter{table}
\centering

\begin{tabularx}{5.5cm}{X|YYY}
         \toprule $c_i$ & $\alpha$ & $\beta_1$ & $\beta_2$  \\ \midrule
         $c_1$ & $0.01$ & $0.9$ & $0.999$ \\ 
         $c_2$ & $0.5$ & $0.9$ & $0.999$ \\ 
         $c_3$ & $0.05$ & $0.9$ & $0.999$ \\ 
         $c_4$ & $0.5$ & $0.9$ & $0.999$ \\ \bottomrule
\end{tabularx}

\label{tab:tab_6}
\end{table}

\noindent {\bf Tab. 4.} Parameters of Adam used for local optimization during the post-processing for the optimization of test functions $c_2$, $c_3$, and $c_4$.
\begin{table}[!ht]
\refstepcounter{table}
\centering

\begin{tabularx}{5.5cm}{X|YYY}
         \toprule $c_i$ & $\alpha$ & $\beta_1$ & $\beta_2$  \\ \midrule
         $c_1$ & $0.001$ & $0.9$ & $0.999$ \\ 
         $c_2$ & $0.5$ & $0.9$ & $0.999$ \\ 
         $c_3$ & $0.05$ & $0.9$ & $0.999$ \\ 
         $c_4$ & $0.5$ & $0.9$ & $0.999$ \\ \bottomrule
\end{tabularx}

\label{tab:tab_3}
\end{table}
\clearpage

\end{document}